\documentclass[10pt,twocolumn,letterpaper]{article}

\usepackage{iccv}
\usepackage{times}
\usepackage{epsfig}
\usepackage{graphicx}
\usepackage{amsmath}
\usepackage{amssymb}
\usepackage{float}
\usepackage{booktabs}
\usepackage{multirow}
\usepackage[dvipsnames,table,xcdraw]{xcolor}

\usepackage{times}  
\usepackage{helvet}  
\usepackage{courier}  
\usepackage[hyphens]{url}  
\usepackage{caption} 
\usepackage{textcomp}
\usepackage{stfloats}
\usepackage{url}
\usepackage{verbatim}
\usepackage{graphicx}
\usepackage{cite}
\usepackage{booktabs}
\usepackage{tabularx}
\usepackage{amsfonts,amssymb}
\usepackage{multirow}
\usepackage{bbding}
\usepackage{fontenc}
\usepackage{amsmath}
\usepackage{array}
\usepackage{xcolor}
\usepackage{colortbl}
\usepackage[caption=false]{subfig}
\usepackage{caption}
\usepackage{wasysym}
\usepackage[pagebackref=true,breaklinks=true,letterpaper=true,colorlinks,bookmarks=false]{hyperref}
\definecolor{bblue}{rgb}{0,150,230}
\definecolor{lightgray}{gray}{.96}
\definecolor{myy}{RGB}{126,95,0}
\definecolor{ggray}{RGB}{127,127,127}
\definecolor{mygreen}{RGB}{93,173,85}
\definecolor{myred}{RGB}{240,16,89}
\definecolor{myblue}{RGB}{0,114,188}
\definecolor{darkgreen}{rgb}{0.0, 0.5, 0.0}
\definecolor{demphcolor}{RGB}{100,100,100}

\newcommand{\sgrouptablestyle}[2]{\setlength{\tabcolsep}{#1}\renewcommand{\arraystretch}{#2}\centering}
\newcolumntype{d}[1]{>{\raggedright\arraybackslash}p{#1pt}}
\newcolumntype{e}[1]{>{\raggedleft\arraybackslash}p{#1pt}}
\newcommand{\hlg}[1]{\textcolor{mygreen}{#1}}
\newcommand{\bbetter}[4]{
    \sgrouptablestyle{1pt}{1}
    \begin{tabular}{e{#1}d{#2}}
    {#3} &
    {\fontsize{6.5pt}{1em}\selectfont \hlg{\textbf{$\uparrow$#4}}}
    \end{tabular}
}

\iccvfinalcopy 

\def\iccvPaperID{3986} 

\definecolor{mygray}{gray}{.92}

\makeatletter
\newcommand{\thickhline}{%
    \noalign {\ifnum 0=`}\fi \hrule height 0.8pt
    \futurelet \reserved@a \@xhline
}
\makeatother
\ificcvfinal\fi

\begin{document}

\title{\vspace{-1cm}Star-Net: Improving Single Image Desnowing Model With More Efficient Connection and Diverse Feature Interaction}

\author{
Jiawei Mao$^\dag$
\quad Yuanqi Chang$^\dag$  \quad Xuesong Yin{\thanks{Corresponding author.$^\dag$Equal contribution.}} \quad Binling Nie  \\ 
School of Media and Design, Hangzhou Dianzi University, Hangzhou, China \qquad \\
{\tt\small\{jiaweima0,yuanqichang,yinxs,binlingnie\}@hdu.edu.cn }\\
}
\maketitle
\ificcvfinal\thispagestyle{empty}\fi

\begin{abstract}
Compared to other severe weather image restoration tasks, single image desnowing is a more challenging task. 
This is mainly due to the diversity and irregularity of snow shape, which makes it extremely difficult to restore images in snowy scenes. 
Moreover, snow particles also have a veiling effect similar to haze or mist. 
Although current works can effectively remove snow particles with various shapes, they also bring distortion to the restored image. 
To address these issues, we propose a novel single image desnowing network called Star-Net. 
First, we design a Star type Skip Connection (SSC) to establish information channels for all different scale features, which can deal with the complex shape of snow particles.
Second, we present a Multi-Stage Interactive Transformer (MIT) as the base module of Star-Net, which is designed to better understand snow particle shapes and to address image distortion by explicitly modeling a variety of important image recovery features. 
Finally, we propose a Degenerate Filter Module (DFM) to filter the snow particle and snow fog residual in the SSC on the spatial and channel domains. 
Extensive experiments show that our Star-Net achieves state-of-the-art snow removal performances on three standard snow removal datasets and retains the original sharpness of the images.
\end{abstract}

\section{Introduction}

The main purpose of single image snow removal is to remove snow particles in the foreground of the image and snow fog in the background. 
Snow particles can obscure the foreground of the image, which affects the machine's observation of the foreground object. 
Snow fog adds a foggy or hazy effect to the image which makes it blurry. 
Therefore, the performance of the algorithm is reduced when performing high-level vision tasks such as image classification, object detection and semantic segmentation in snowy scenes. 
This appears to be particularly important for single image desnowing tasks.

\begin{figure}[t]
\centering
\setlength{\belowcaptionskip}{-0.8cm}
\includegraphics[width=1\linewidth]{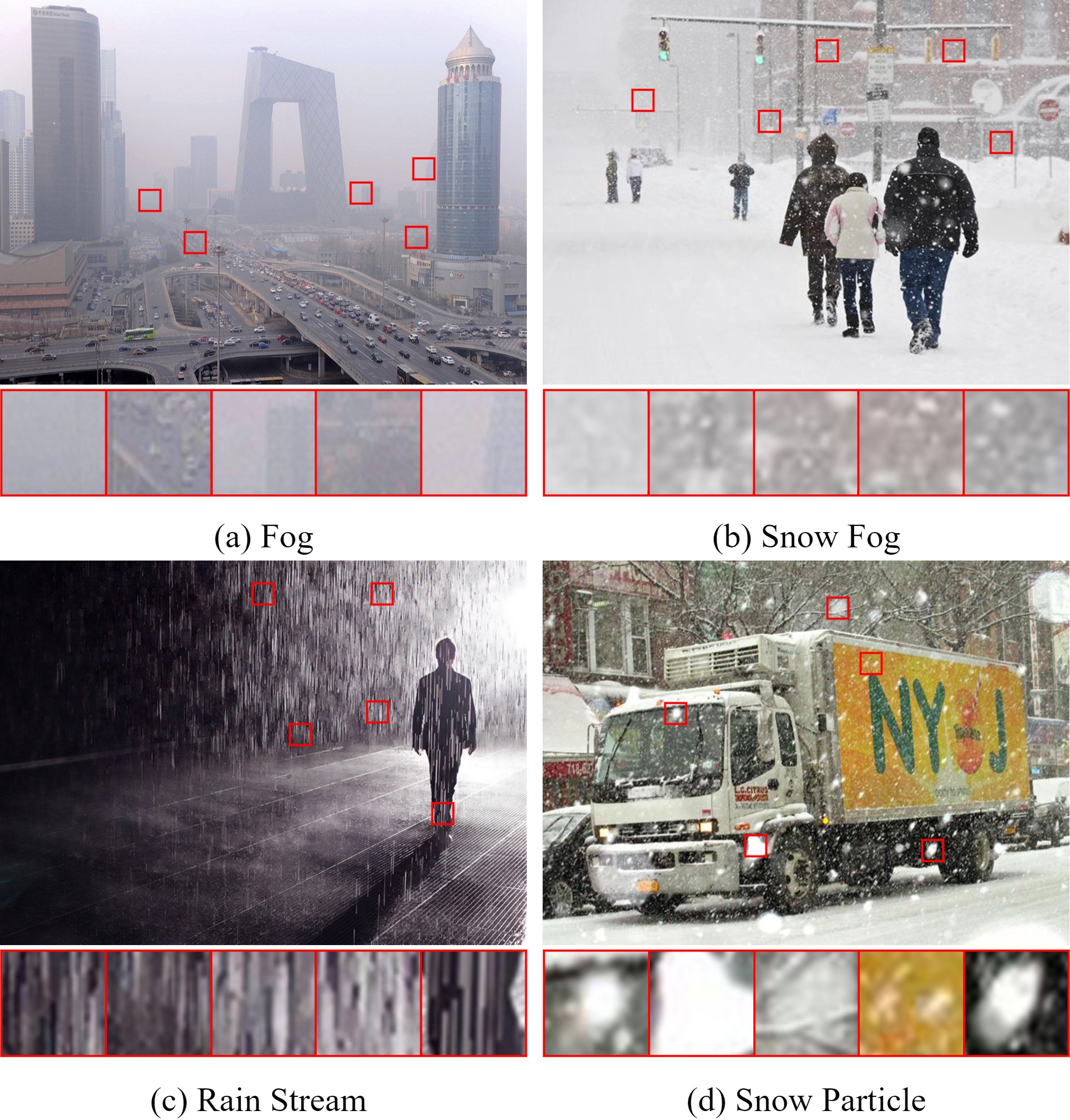}
\caption{Comparison of various severe weather image with snow images. 
According to the rectangular border area shown, the \textbf{snow particle} show a complex as well as a more irregular shape compared to the \textbf{rain stream}. 
Compared to \textbf{fog}, \textbf{snow fog} interspersed with snow particle makes the image more difficult to identify. Please zoom in to see the details better.}
\label{fig1}
\end{figure}

Single image snow removal is more difficult than other severe weather image recovery tasks. 
As can be seen in Figure~\ref{fig1}, the shape of the snow particles shows more irregularity and diversity compared to the rain streak. 
The combination of snow fog and snow particles causes the image more difficult to recognize compared to haze scene. 
Many methods have been proposed to meet the challenge posed by the single image desnowing task.

\begin{figure*}[h]
\centering
\setlength{\abovecaptionskip}{0.2cm}
\setlength{\belowcaptionskip}{-0.6cm}
\includegraphics[width=1\linewidth]{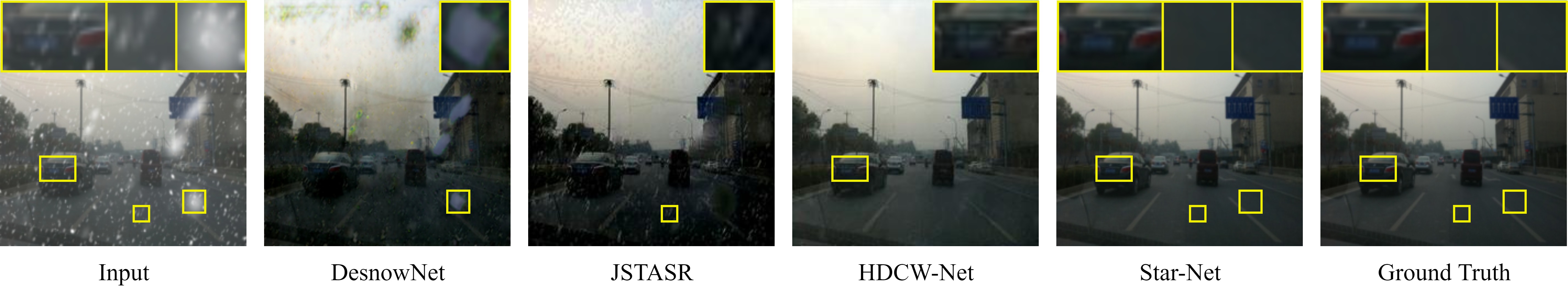}
\caption{Issues with the current single image desnowing algorithms. As shown in the rectangular area in the figure, DesnowNet \cite{liu2018desnownet} and JSTASR \cite{chen2020jstasr} can't completely remove snow particles of different sizes. 
Compared with ground truth, HDCW-Net \cite{chen2021all} suffers from image distortion. 
Our Star-Net provides a good solution to the above issues. Please zoom in to see the details better.}
\label{fig2}
\end{figure*}

Overall, these methods can be broadly divided into three categories: manual feature-based snow removal methods \cite{xu2012improved,zheng2013single,pei2014removing,wang2017hierarchical,yu2014content,yan2019snow,barnum2007spatio}, 
convolutional hierarchy-based snow removal methods \cite{liu2018desnownet,chen2020jstasr,chen2021all,li2020all,li2019stacked,li2019single} and attention-based snow removal methods \cite{valanarasu2022transweather,chen2022learning}. 
The manual feature-based snow removal methods fully utilize a prior knowledge to strip snow particles from the image background. 
However, due to the complex and variable shape of snow particles, these methods are poor in generalization to other snow scenarios. 
To solve the above issue, Liu \etal \cite{liu2018desnownet} proposed the first convolution-based snow removal algorithm. 
Chen \etal \cite{chen2020jstasr} proposed a convolutional snow removal algorithm based on snow particle shape and transparency, which performs the snow removal task by image recovery. 
And they also used the properties of dual-tree wavelet transform in HDCW-Net \cite{chen2021all} to better learn the complex shape of snow particles. 
To learn the global structure of severe weather images, Li \etal \cite{li2020all} designed an attention-based unified severe weather recovery transformer network. 
And this network performs well in a single image desnowing task. 
In addition, Tu \etal \cite{tu2022maxim} designed a general-purpose vision backbone based on a multi-axis multi-layer perceptron (MLP) architecture that achieved promising results on a variety of severe weather image recovery tasks. 
Despite the impressive performance achieved by these methods, there is still room for improvement in these solutions.

Overall, there are two issues that need to be addressed for the single image desnowing task nowadays (see Figure~\ref{fig2}).

\textbf{(i) Diversity and irregularity of snow particles shape: }Snow particles exist in varying sizes and irregular shapes. This poses a huge challenge for single image desnowing work. 
As can be seen in Figure~\ref{fig2}, most single image desnowing methods only consider snow particle at a single scale and are not able to do both removal for large snow particles and small snow particles. This can be solved with the help of multi scale features.

\textbf{(ii) Distortion after single image desnowing: }The sharpness of the snow removal image is also an important indicator to evaluate the quality of the single image desnowing model. 
In Figure~\ref{fig2}, although HDCW-Net is better at removing snow particles of different scales, there is a serious image distortion problem in the image after desnowing. 
Enhanced local modeling helps image after desnowing to maintain the sharpness of the original image.

Therefore, in this paper, we propose a novel single image desnowing transformer network to address the above issues called Star-Net. 
Firstly, we design a Star type Skip Connection (SSC) for Star-Net. 
To enable effective interaction of features at different scales, SSC establishes information channels for different levels of encoder and decoder modules as well as the Degenerate Filter Module (DFM) we designed. 
This introduce a multi scale characteristic for Star-Net to learn the complex and variable shapes of snow particles. 
Secondly, we design a Multi-Stage Interactive Transformer (MIT) as the basic unit of Star-Net, which explicitly models several important recovery features through multiple attention interactions and multi scale deep convolution. 
This further makes Star-Net focus on multi scale feature modeling capabilities and introduce rich local details for solving image distortion issue. 
In addition, we find snow particles and snow fog residuals exist in the SSC (see Figure~\ref{fig9}). 
Thus, we propose a DFM for SSC by combing spatial attention module and channel gating module. 

Numerous experiments show that Star-Net achieves state-of-the-art (SOTA) performance on several standard snow removal datasets. 
Specifically, Star-Net achieves a PSNR of 44.04dB and SSIM of 0.99 on CSD \cite{chen2021all}. 
In addition, Star-Net obtains PSNR gains of 0.14dB and 0.12dB on the SRRS \cite{chen2020jstasr} and Snow100K \cite{liu2018desnownet} datasets, respectively. 

\section{Related Work}

\subsection{Single image snow removal}

Liu \etal \cite{liu2018desnownet} proposed the first deep learning-based snow removal network and the first large benchmark dataset for snow removal. 
This method removes the image snow information step-by-step by labeling the snowflakes with attributes and a multi-stage network. 
Li \etal \cite{li2019stacked} proposed a snow removal algorithm that combines the physical imaging model of snowflakes and DenseNet \cite{huang2017densely}. 
Li \etal \cite{li2019single} designed an image desnowing algorithm based on generative adversarial networks. 
Chen \etal \cite{chen2020jstasr} achieved image desnowing with different snow concentrations by adding transparency perception to the snow scene images, allowing the model to obtain better results in both non-transparent and transparent snowy scenes. 
Additionally, they also added veiling effects to snow images to synthesize the SRRS desnowing dataset. Li \etal \cite{li2020all} designed a general network for severe weather such as rain, snow, and fog based on adversarial learning strategy. 
By combining dual-tree wavelet transform and convolutional neural network, Chen \cite{chen2021all} designed a hierarchical progressive structure of snow removal network. 
These single image desnowing methods based on convolutional neural networks exhibit good generalization capabilities.

\subsection{Vision Transformer in low-level visual tasks}

Recently, ViT \cite{dosovitskiy2020image} has made great progress in computer high-level vision tasks such as, image classification \cite{zhang2022parc,lanchantin2021general,chen2021crossvit,liu2021swin}, 
object detection \cite{carion2020end,chen2021pix2seq,zhu2020deformable,wang2021pnp,roh2021sparse,wang2022anchor} and semantic segmentation \cite{zheng2021rethinking,chen2021transunet,xie2021segformer,strudel2021segmenter,hoyer2022daformer,he2022swin}. 
Compared to CNN, the transformer has powerful long-range modeling capability, which greatly helps the model to learn the global information. 
Meanwhile, more and more attention has been paid to the application of the transformer in low-level visual tasks. 
Chen \etal \cite{chen2021pre} first introduced transformer to low-level vision by pre-training on ImageNet \cite{deng2009imagenet} and fine-tuning it in specific datasets to achieve tasks such as image denoising, deraining, and super-resolution. 
Through combining transformer block and U-Net, Wang \etal \cite{wang2022uformer} used deep separable convolution as feedforword network to accomplish modeling of contextual information for image denoising, rain removal, and deblurring. 
Liang \etal \cite{liang2021swinir} developed an image restoration network based on Swin Transformer\cite{liu2021swin}. Nevertheless, the above work has a very large computational overhead due to the long range modeling. 
To solve this problem, Zamir \etal \cite{zamir2022restormer} designed a transformer for the channel domain to replace original computation for the spatial domain, which performs remarkable in tasks such as deraining, denoising and deblurring. 
Valanarasu \etal \cite{valanarasu2022transweather} proposed an end-to-end network which enabled the recovery of multiple severe weather images together. 
All these transformer-based methods show SOTA performance in low-level vision tasks.

\section{Method}

As shown in Figure~\ref{fig3}, Star-Net is mainly composed of three key designs: Star type Skip Connection, Multi-Stage Interactive Transformer and Degenerate Filter Module. 
SSC introduces multi scale characteristics to Star-Net. 
MIT implements explicit modeling of several important recovery features. DFM helps filter snow particles and snow fog degradation residuals in SSC.

\subsection{Star type Skip Connection}

\begin{figure}[h]
   \centering
   \setlength{\belowcaptionskip}{-0.4cm}
   \includegraphics[width=1\linewidth]{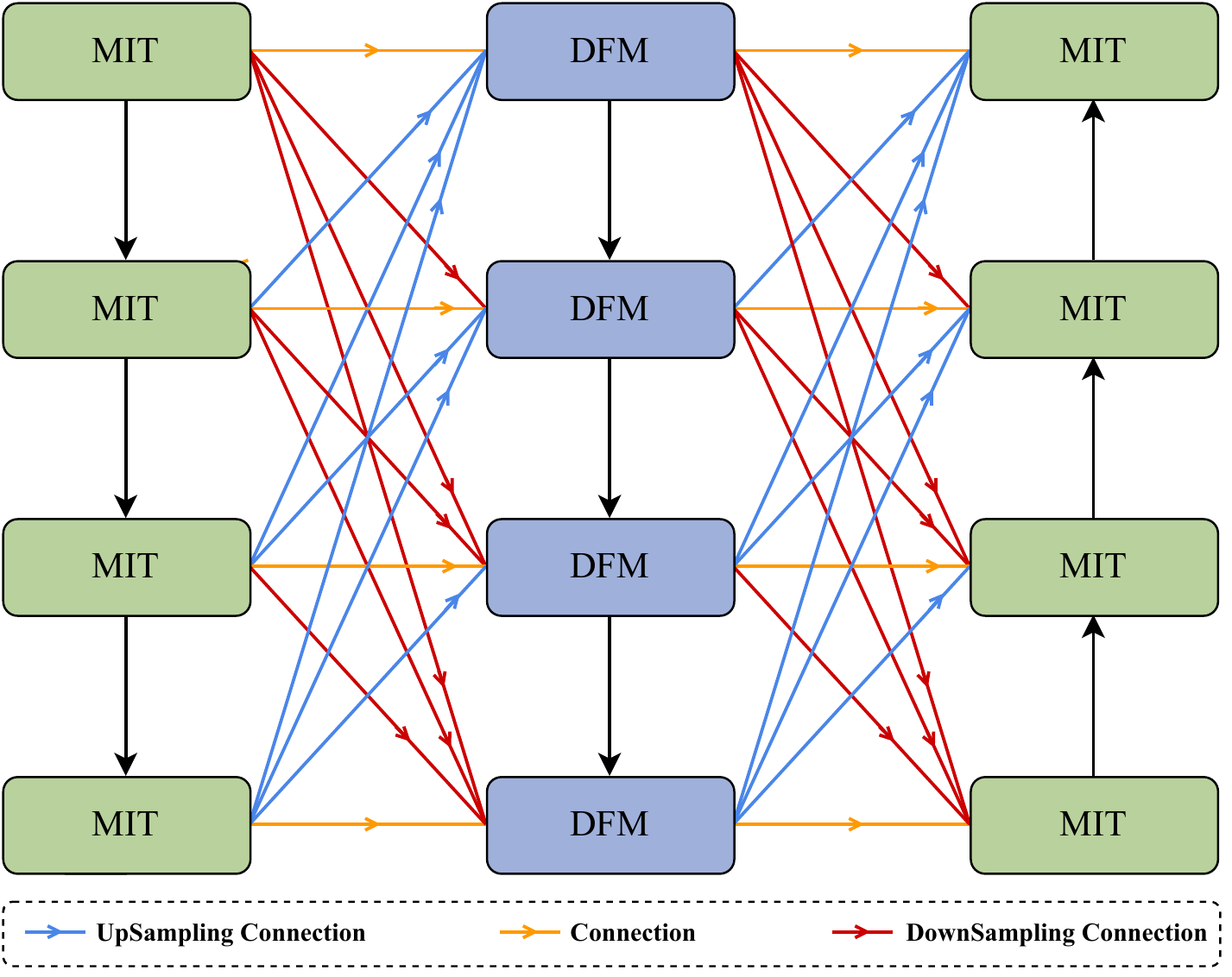}
   \caption{The Star-Net pipeline. Star-Net consists of three core designs: SSC, MIT, and DFM.}
   \label{fig3}
   \end{figure}

   \begin{figure*}[h]
\centering
\setlength{\belowcaptionskip}{-0.6cm}
\includegraphics[width=0.82\linewidth]{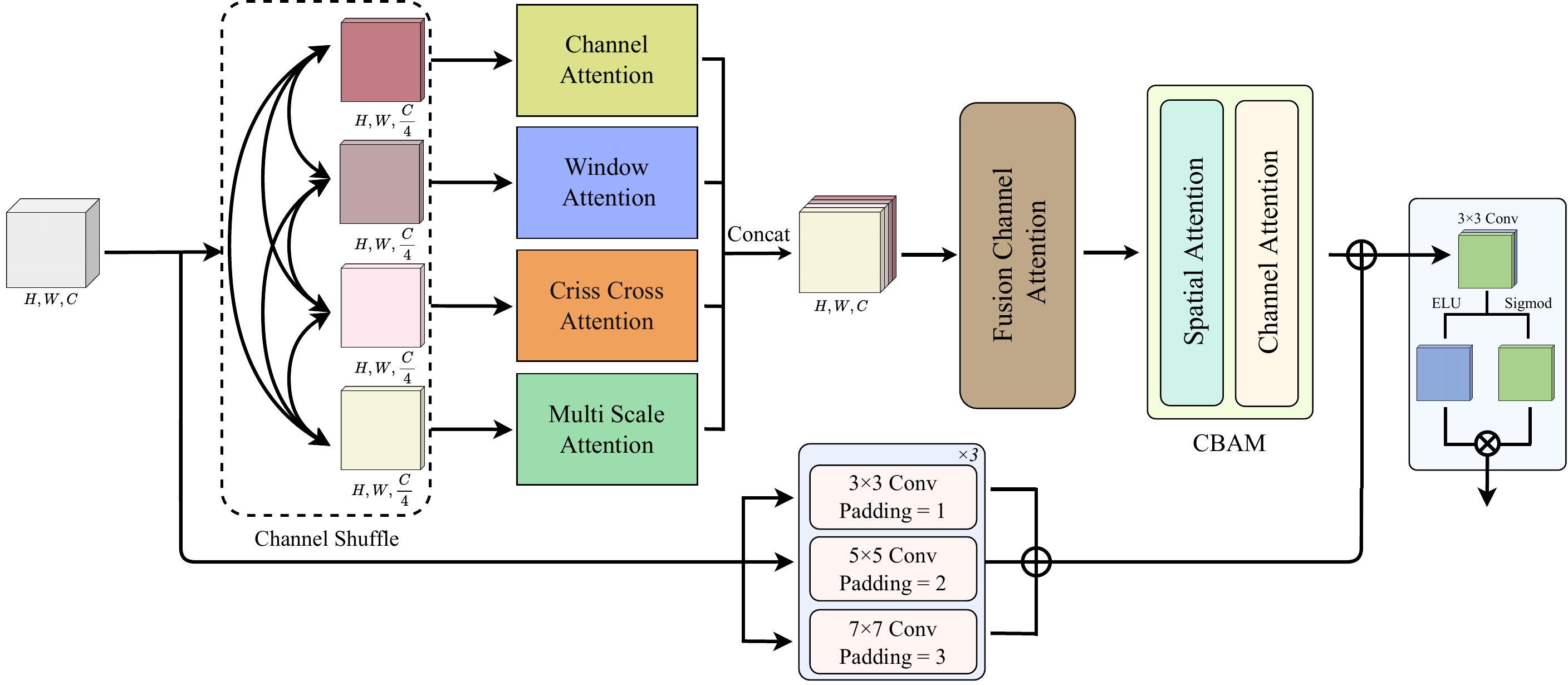}
\caption{Detailed internal structure of MIT. MIT mainly contains three core designs of \textbf{Multi Stage Attention Mechanisms} composed of multiple attention mechanisms, 
\textbf{Multi Scale Deep Convolution} with different scale convolution kernels, and \textbf{Channel Gating Network.}}
\label{fig4}
\end{figure*}
The Star-Net backbone (Figure~\ref{fig3}) adopts U-Net's encoder-decoder design guidelines. 
We argue that the multi scale structural design facilitates the model to learn the complex and variable shapes of snow particles compared to single scale skip connection in U-Net \cite{ronneberger2015u}. 
Therefore, we create connections for each unit of Star-Net in each layer to units that are not on the same layer. We call this star type skip connection method SSC.

We mainly analyze SSC in three information flow directions: from encoder module to DFM, from DFM to decoder module, and from DFM to DFM. 

First, we define three operations of SSC: upsampling connection, downsampling connection, and connection. 
Given the inputs ${\text{F}_0},...,{\text{F}_{n}}$ ($n$ denotes the number of network layers), we define the \textbf{downsampling connection} for ${i^{th}}$ layer as:
\begin{equation}  
\setlength{\abovedisplayskip}{0pt}
\setlength{\belowdisplayskip}{0pt}
\text{downsamplin}{{\text{g}}_{\text{i}}}({\text{F}}) = \sum\limits_{{j} = 0}^{{i} - 1} {{\text{con}}{{\text{v}}_{1}}{\text{(con}}{{\text{v}}_3}({{\text{F}}_{\text{j}}}))},
\label{eq1}
\end{equation}
where ${\text{con}}{{\text{v}}_\text{3}}( \cdot )$ denotes a downsampling operation with $3 \times 3$ convolution kernels used to adjust different features to fit the shape of ${i^{th}}$ layer feature shape.
${\text{con}}{{\text{v}}_1}( \cdot )$ represents the alignment of the channel dimension to the ${i^{th}}$ layer feature with a $1 \times 1$ convolution kernel.

The \textbf{upsampling connection} for ${i^{th}}$ layer is computed as:
\begin{equation}
  \setlength{\abovedisplayskip}{0pt}
\setlength{\belowdisplayskip}{0pt}
\text{upsampling}_{\text{i}}({\text{F}}) = \sum\limits_{j = {i} + 1}^{n} {{\text{con}}{{\text{v}}_{\text{1}}}{\text{(decon}}{{\text{v}}_3}({{\text{F}}_{\text{j}}}))},
\label{eq2}
\end{equation}
where ${\text{decon}}{{\text{v}}_3}( \cdot )$ denotes a upsampling operation with $3\times3$ convolution kernels for adjusting different features to fit the shape of ${i^{th}}$ layer feature shape.

The ${i^{th}}$ layer \textbf{connection} can be expressed as:
\begin{equation}
  \setlength{\abovedisplayskip}{5pt}
\setlength{\belowdisplayskip}{5pt}
{\text{connec}}{{\text{t}}_{\text{i}}}({\text{F}}) = {\text{con}}{{\text{v}}_1}({{\text{F}}_{\text{i}}}),
\label{eq3}
\end{equation}

\noindent\textbf{From Encoder Module to DFM. }Formally, given the encoder outputs for each layer ${\text{e}_i},{i \in [0,{n}}]$, for the DFM at each layer, it receives the multi scale information streams from each layer encoder module via $\text{SSC}$. This can be indicated as:
\begin{equation}
\begin{split}
&{\text{SS}}{{\text{C}}_{\text{i}}}({\text{e}}) = {\text{connec}}{{\text{t}}_{\text{i}}}({\text{e}})\\
&+{\text{upsamplin}}{{\text{g}}_{\text{i}}}({\text{e}}) + {\text{downsamplin}}{{\text{g}}_{\text{i}}}({\text{e}}),{\text{i}} \in [0,{n}],
\end{split}
\label{eq4}
\end{equation}

\noindent\textbf{From DFM to Decoder Module. }Specially, given the DFM outputs for each layer ${{\text{d}}_i}$, for the decoder module at each layer, 
it receives the multi scale information streams from each layer DFM via SSC. This can be shown as: 
\begin{equation}
\begin{split}
&{\text{SS}}{{\text{C}}_{\text{i}}}({\text{d}}) = {\text{connec}}{{\text{t}}_{\text{i}}}({\text{d}}) \\
&+ {\text{upsamplin}}{{\text{g}}_{\text{i}}}({\text{d}}) + {\text{downsamplin}}{{\text{g}}_{\text{i}}}({\text{d}}),{\text{i}} \in [0,{n}],
\end{split}
\label{eq5}
\end{equation}

\noindent\textbf{From DFM to DFM. }Suppose the output of DFM at each layer are ${{\text{m}}_0},...,{{\text{m}}_{n}}$. 
For each layer of DFM, it receives the output from the previous layer of DFM via SSC.
\begin{equation}
{\text{SS}}{{\text{C}}_{\text{i}}}({\text{m}}) = {\text{conv}}({{\text{m}}_{{\text{i-1}} }}),{\text{i}} \in [1,{n}],
\label{eq6}
\end{equation}

For different layer of DFM output $\text{m}$, we adopt different size convolution kernel ${\text{conv}}( \cdot )$ for scaling (For example, we use a convolutional kernel of size $7$ for $\text{m}_{0}$ and a convolutional kernel of size $5$ for $\text{m}_{1}$).
The motivation for this branch design is to further enhance the multi scale characteristic of Star-Net and to integrate all multi scale information fused by DFM into the bottleneck of Star-Net.

\subsection{Multi-Stage Interactive Transformer}
To address the image distortion issue that occurs in the single image desnowing in Figure~\ref{fig2} and to enhance the learning ability of Star-Net for snow particles, we carefully thought about the design of the Star-Net basic unit.

For the image distortion issue, we solve it by enhancing the local modeling capability in parallel from both the window attention (WA) mechanism \cite{liu2021swin} and convolution directions. 
To further strengthen Star-Net's ability to learn the variable shape of snow particles, we also introduce the multi scale attention (MSA) mechanism \cite{ren2022shunted} in the basic unit. 
Considering the problem of large scale snow particle removal, we use global features to assist the solution. 
This leads us to introduce the criss cross attention (CCA) mechanism \cite{huang2019ccnet} that is computationally low and has global modeling capabilities. 
Considering that feature channels are rich in image recovery information \cite{zamir2022restormer} and the channel attention (CA) mechanism has the advantage of small computational cost, we introduce it into the basic unit of Star-Net. 
In addition, we replace the vanilla feedforward network with a convolutional gating network to further enhance the snow particle removal capability. 
We call the unit combined with the above design as Multi-Stage Interactive Transformer (Figure~\ref{fig4}).
Next, we explain in detail several key designs of MIT. MIT is composed of multi stage attention mechanisms, multi scale deep convolution, and convolutional gating network.

\noindent\textbf{Multi Stage Attention Mechanisms. }Given an input feature ${\text{G}} \in {{\text{R}}^{{{H}} \times {{W}} \times {{C}}}}$, 
we first perform a channel shuffle operation on it and divide it into ${{\text{G}}_{\text{i}}} \in {{\text{R}}^{{{H}} \times {{W}} \times \frac{{{C}}}{{4}}}},{i} \in [1,...,4]$ according to the channel dimension. 
The channel shuffle operation ensures the disorder of each channel feature and avoids the single attention modeling always executed for the same channel feature. Overall, multi stage attention can be modeled as:
\begin{equation}
  \setlength{\abovedisplayskip}{3pt}
  \setlength{\belowdisplayskip}{3pt}
\begin{split}
&{\text{G}_{1}} = {\text{WA(}}{\text{G}_{1}}{\text{), }}{{\text{G}}_{\text{2}}} = {\text{MSA(}}{{\text{G}}_{\text{2}}}{\text{), }}\\
&{\text{G}_{3}} = {\text{CCA(}}{\text{G}_{3}}{\text{), }}{\text{G}_{4}} = {\text{CA(}}{\text{G}_{4}}{\text{),}}\\
&{\text{G}} = {\text{[}}{{\text{G}}_{\text{1}}}{\text{,}}{{\text{G}}_{\text{2}}}{\text{,}}{{\text{G}}_{\text{3}}}{\text{,}}{{\text{G}}_{\text{4}}}{\text{],}}\\
&{\text{T}} = {\text{CBAM(CA(G)),}}
\end{split}
\label{eq7}
\end{equation}
where $[ \cdot ]$ denotes the concat operation and ${\text{CBAM}} (\cdot ) $ is the convolutional block attention module \cite{woo2018cbam}. 
The reason we introduce lightweight CBAM for MIT and $\text{CA}$ for $\text{G}$ is to integrate image recovery features modeled by different attention mechanisms.

\noindent\textbf{Multi Scale Deep Convolution. }For the input feature $\text{G}$, we perform multi scale modeling and local modeling from a convolutional perspective. 
The multi scale deep convolution can be computed as:
\begin{equation}
  \setlength{\abovedisplayskip}{3pt}
  \setlength{\belowdisplayskip}{3pt}
\begin{split}
&{{\text{H}}_{\text{1}}} = {\text{con}}{{\text{v}}_{\text{3}}}({\text{con}}{{\text{v}}_{\text{3}}}({\text{con}}{{\text{v}}_{\text{3}}}({\text{G}}))),\\
&{{\text{H}}_{\text{2}}} = {\text{con}}{{\text{v}}_{\text{5}}}({\text{con}}{{\text{v}}_{\text{5}}}({\text{con}}{{\text{v}}_{\text{5}}}({\text{G}}))),\\
&{{\text{H}}_{\text{3}}} = {\text{con}}{{\text{v}}_{\text{7}}}({\text{con}}{{\text{v}}_{\text{7}}}({\text{con}}{{\text{v}}_{\text{7}}}({\text{G}}))),\\
&{\text{H}} = {{\text{H}}_{\text{1}}} + {{\text{H}}_{\text{2}}} + {{\text{H}}_{\text{3}}},
\end{split}
\label{eq8}
\end{equation}
where ${\text{con}}{{\text{v}}_\text{5}}( \cdot )$ and ${\text{con}}{{\text{v}}_\text{7}}( \cdot )$ represent convolutions with convolution kernels of $5$ and $7$, respectively.

\noindent\textbf{Convolutional Gating Network. }We use a convolutional gating network as a feedforward network for MIT to further remove snow particle residuals, which can be expressed as:
\begin{equation}
  \setlength{\abovedisplayskip}{3pt}
  \setlength{\belowdisplayskip}{3pt}
\begin{split}
&{\text{O}} = {\text{H}} + {\text{T,}}\\
&{\text{[}}{{\text{O}}_{\text{1}}},{{\text{O}}_{\text{2}}}] = \varphi ({\text{con}}{{\text{v}}_{\text{3}}}({\text{O}})){\text{,}}\\
&{{\text{O}}_{\text{1}}} = {\text{sigmoid}}({{\text{O}}_{\text{1}}}),{{\text{O}}_{\text{2}}} = {\text{ELU}}({{\text{O}}_{\text{2}}}),\\
&{\text{O}} = {{\text{O}}_{\text{1}}} \otimes {{\text{O}}_{\text{2}}},
\end{split}
\label{eq9}
\end{equation}
where $\varphi (\cdot)$ denotes the split operation on the channel dimension, ${\text{sigmoid}}( \cdot )$ and ${\text{ELU}}( \cdot {)}$ denotes the sigmoid activation function \cite{han1995influence} and ELU activation function \cite{clevert2015fast}, respectively. 
And $ \otimes $ denotes the Hadamard product of matrices.

\subsection{Degenerate Filter Module}
\begin{figure}[h]
  \centering
  \setlength{\belowcaptionskip}{-0.4cm}
  \includegraphics[width=1\linewidth]{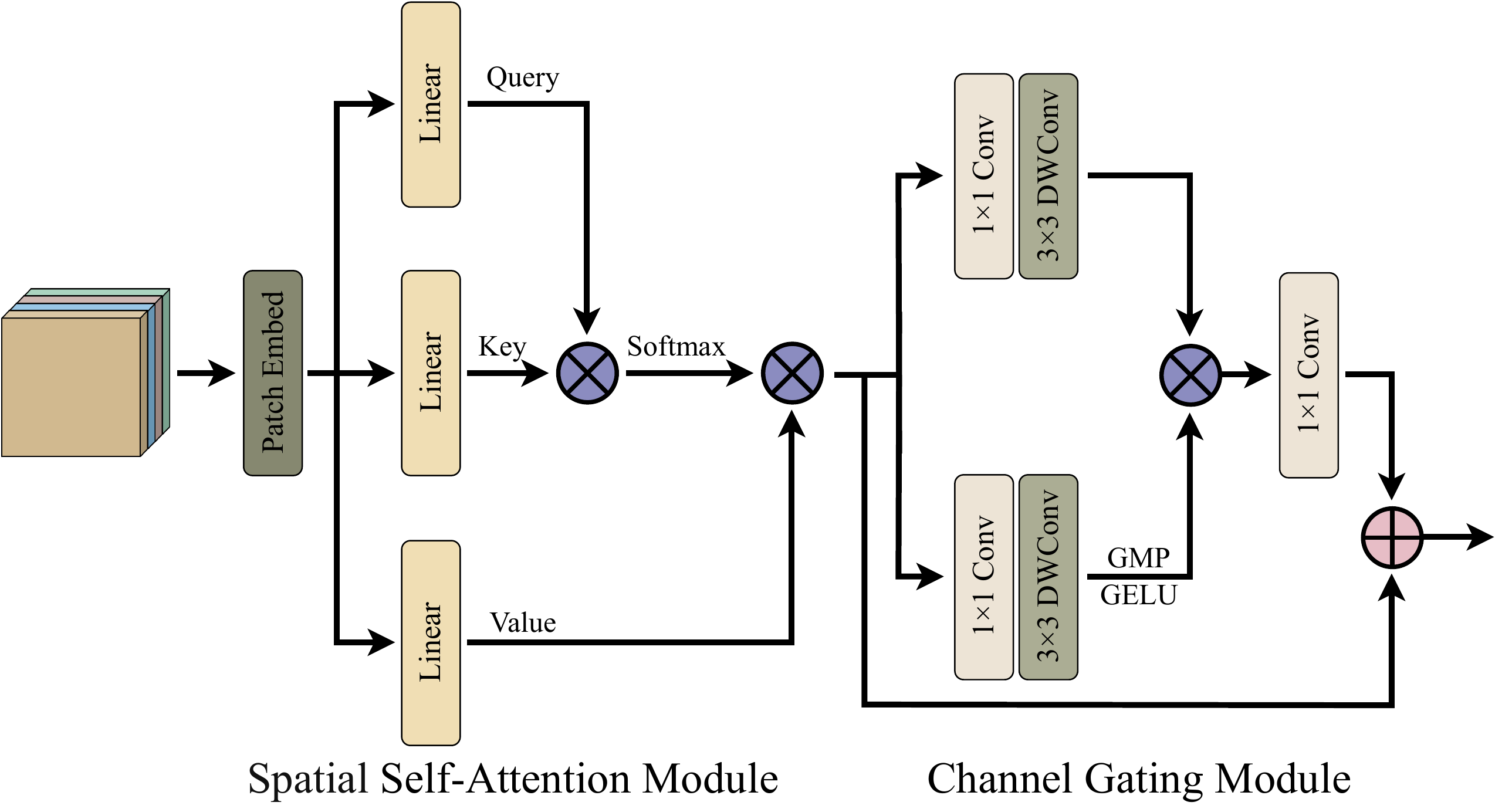}
  \caption{The architecture of DFM for snow particle and snow fog filtering.}
  \label{fig5}
  \end{figure}
To prevent the SSC from transmitting snow particles and snow fog residuals to Star-Net's decoder module, we design the degradation filter module (Figure~\ref{fig5}) in the SSC. 
In addition to the possible presence of snow particles and snow fog residuals in the spatial domain, they may also be present on the channel domain. 
Thus, our DFM takes into account both spatial domain and channel domain snow particles and snow fog residuals.

Specifically, DFM consists of a spatial self-attention module and channel gating module. The spatial self-attention module mainly deals with snow residuals in the spatial domain. 
The channel gating module mainly deals with snow residuals in the channel domain. 

\noindent\textbf{Spatial Self-Attention Module. }Given embedding patches ${\text{O}}$, spatial self-attention module strips snow particles and snow fog residuals from the image by interacting with each patch on the space.
The spatial self-attention module is modeled as:
\begin{equation}
  \setlength{\abovedisplayskip}{3pt}
  \setlength{\belowdisplayskip}{3pt}
\begin{split}
      &{\text{Q}} = {{\text{W}}_{\text{Q}}}{\text{O}},{\text{K}} = {{\text{W}}_{\text{K}}}{\text{O}}, {\text{V}} = {{\text{W}}_{\text{V}}}{\text{O,}}\\
      &{\text{O}} = {\text{softmax}}(\frac{{{\text{Q}}{{\text{K}}^{\text{T}}}}}{{\sqrt {\text{d}} }}){\text{V}},
\end{split}
\label{eq10}
\end{equation}
where ${{\text{W}}_{\text{Q}}},{{\text{W}}_{\text{K}}},{{\text{W}}_{\text{V}}}$ represent the mapping matrices, respectively. ${\text{softmax}}( \cdot )$ is the softmax activation function \cite{brown1992class}. \text{d} indicates the embedding dimension. 

\noindent\textbf{Channel Gating Module. }For the spatial self-attention module output, we adopt gating mechanism to control the information flow in each channel. 
This allows DFM to remove as much snow residuals as possible trapped in the channel. The channel gating module can be expressed as follows:
\begin{equation}
  \setlength{\abovedisplayskip}{3pt}
  \setlength{\belowdisplayskip}{3pt}
\begin{split}
      &{{\text{O}}_{\text{1}}} = {\text{con}}{{\text{v}}_1}({\text{dwcon}}{{\text{v}}_3}({\text{O}})),\\
      &{{\text{O}}_{\text{2}}} = {\text{con}}{{\text{v}}_1}({\text{dwcon}}{{\text{v}}_3}({\text{O}})),\\
      &{\text{O}} = {\text{con}}{{\text{v}}_1}({\text{GMP(GELU(}}{{\text{O}}_{\text{1}}}{\text{))}} \odot {{\text{O}}_{\text{2}}}),
\end{split}
\label{eq11}
\end{equation}
where ${\text{dwcon}}{{\text{v}}_3}(\cdot )$ denotes a deep convolution with a convolution kernel size of $3$. ${\text{GELU}} (\cdot )$ represents the GELU activation function \cite{hendrycks2016bridging} and $\odot $ means the dot product of elements. 
${\text{GMP}}( \cdot )$ indicates that the global max pooling operation is used to refine the channel information.

\subsection{Loss Function}
The loss function of Star-Net is divided into two parts, the smooth L1-loss \cite{girshick2015fast} between the predicted and ground truth and the perceptual loss of the features extracted by the VGG16 \cite{simonyan2014very} network. 
First, we use the smooth L1-loss based on the correspondence between the predicted (P) and ground truth (G), which can be expressed as:
\begin{equation}
  \setlength{\abovedisplayskip}{3pt}
  \setlength{\belowdisplayskip}{3pt}
{\mathcal{L}_{Smoot{h_{{L_1}}}}} = \left\{ {\begin{array}{ll}
0.5{\theta ^2},&\text{if}\left| \theta  \right| < 1\\
\left| \theta  \right| - 0.5,&\text{otherwise}
\end{array}} \right.
\label{eq12}
\end{equation}
where $\theta  = {\text{P - G}}$. Secondly, the perceptual loss is used by Star-Net to measure the difference between predicted and ground truth features. 
We extract the features in P and G separately by pretraining VGG16 network on ImageNet \cite{deng2009imagenet}, defined as follows: 
\begin{equation}
  \setlength{\abovedisplayskip}{3pt}
  \setlength{\belowdisplayskip}{3pt}
{{\mathcal{L}}_{{perceptual}}} = {\text{MSE}}({\text{VG}}{{\text{G}}_{16}}({\text{P}}),{\text{VG}}{{\text{G}}_{16}}({\text{G}}))
\label{eq13}
\end{equation}
where ${\text{MSE}}( \cdot )$ is mean square error \cite{bauer1999empirical}. In total, our total loss can be expressed as:
\begin{equation}
  \setlength{\abovedisplayskip}{3pt}
  \setlength{\belowdisplayskip}{3pt}
{{\mathcal{L}}_{{total}}} = {{\mathcal{L}}_{{{Smooth}_{{L1}}}}} + {a}{{\mathcal{L}}_{{perceptual}}}
\label{eq14}
\end{equation}
where ${a} = 0.04$ is the hyperparameter of Star-Net used to control the percentage of smooth L1-loss and $\mathcal{L}_{perceptual}$ in the total loss.

  \begin{figure*}[t]
\centering
  \setlength{\abovecaptionskip}{0.1cm}
\setlength{\belowcaptionskip}{-0.3cm}
\includegraphics[width=1\linewidth]{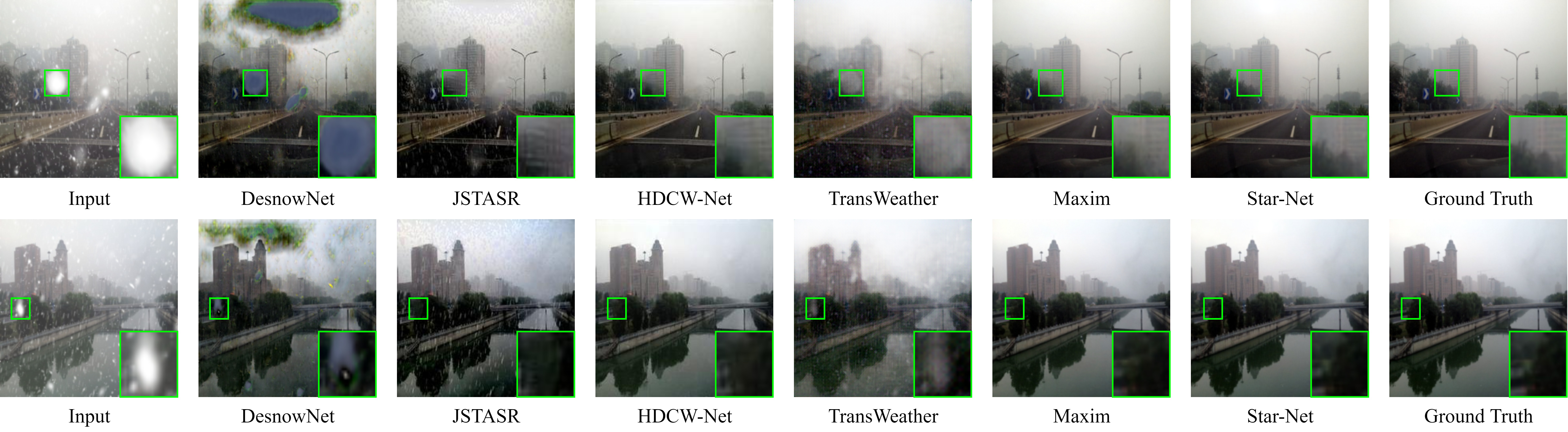}
\caption{Qualitative comparison of Star-Net with several SOTA algorithms for single image desnowing on CSD dataset. Please zoom in to see the details better.}
\label{fig6}
\end{figure*}

\begin{figure*}[t]
\centering
\setlength{\abovecaptionskip}{0.1cm}
\setlength{\belowcaptionskip}{-0.3cm}
\includegraphics[width=1\linewidth]{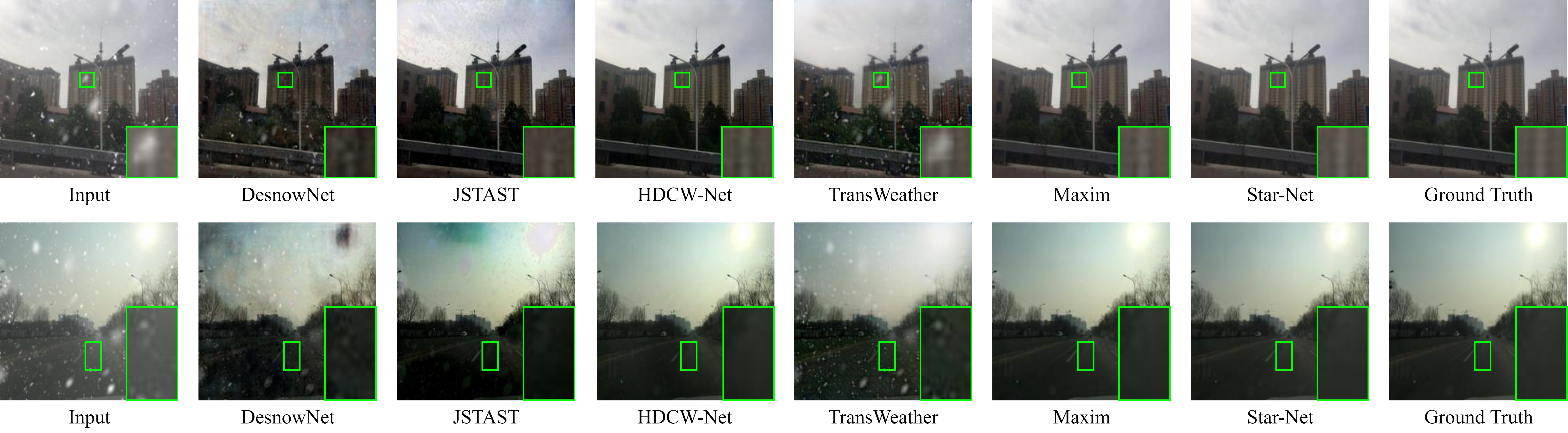}
\caption{Qualitative comparison of Star-Net with several SOTA algorithms for single image desnowing on SRRS dataset. Please zoom in to see the details better.}
\label{fig7}
\end{figure*}

\begin{figure*}[h]
\centering
\setlength{\abovecaptionskip}{0.2cm}
\setlength{\belowcaptionskip}{-0.4cm}
\includegraphics[width=1\linewidth]{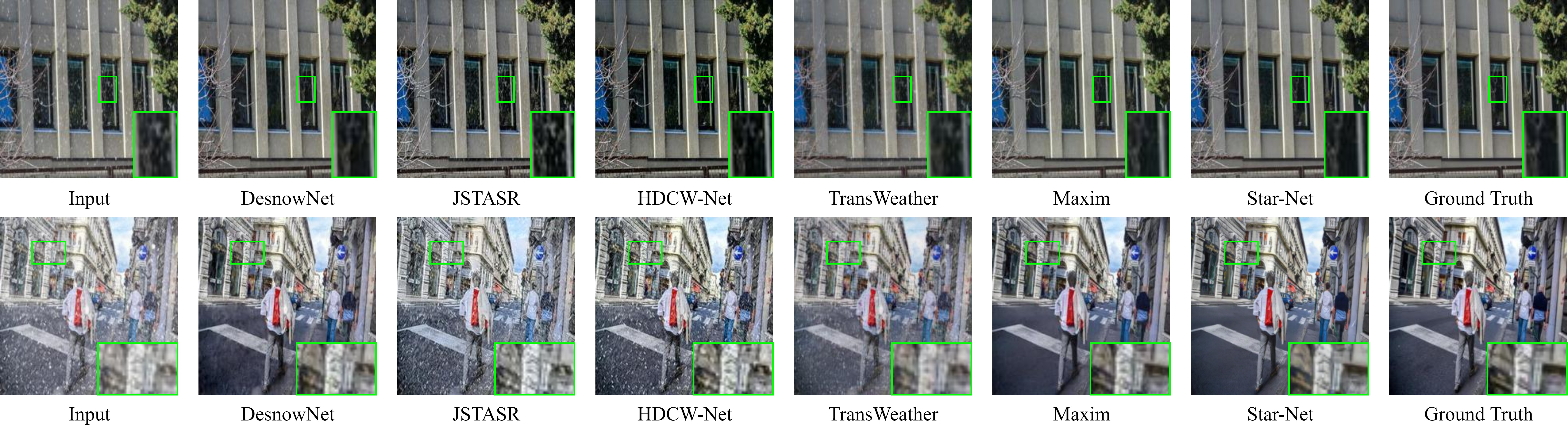}
\caption{Qualitative comparison of Star-Net with several SOTA algorithms for single image desnowing on Snow100K dataset. Please zoom in to see the details better.}
\label{fig8}
\end{figure*}
\section{Experiments}

\subsection{Dataset and Implementation Detail}
\begin{table}[h]
  \centering
  \scriptsize
  \setlength{\abovecaptionskip}{0cm} 
  \setlength{\belowcaptionskip}{-0.4cm}
  \resizebox{8.3cm}{!}{
  \begin{tabular}{c|ccc|c|c}
  \rowcolor{mygray}
  \hline\thickhline
   & \multicolumn{3}{c|}{\text { \textbf{Snow100K} }} & &  \\\cline{2-4}
   \rowcolor{mygray}
  \multirow{-2}{*}{\textbf { Dataset }}             & \text {Snow100K-S}     & \text{Snow100K-M}    & \text{Snow100K-L}  &\multirow{-2}{*}{\textbf{SRRS}}&\multirow{-2}{*}{\textbf{CSD}}\\
  \hline \text {Train set}                                & 2666        &  2667   &  2667 & 8000&8000\\
  \text {Test set}                                         &  666        & 667    &  667 & 2000&2000 \\
  \hline\thickhline
  \end{tabular}}
  \caption{Distribution of the three snow removal benchmark datasets.}
  \label{tab1}
  \end{table}

\begin{table*}[h]
  \centering
  \setlength{\abovecaptionskip}{0.1cm} 
  \setlength{\belowcaptionskip}{-0.4cm}
   \scriptsize
  \renewcommand\arraystretch{1.1}
  \begin{tabular}{c|c|cc|cc|cc}
  \hline\thickhline
  \rowcolor{mygray}
   &  &\multicolumn{2}{c|}{ \textbf{CSD (2000)} } &\multicolumn{2}{c|}{ \textbf{SRRS (2000)} } & \multicolumn{2}{c}{ \textbf{Snow 100K (2000)} }  \\
   \cline{3-8}
   \rowcolor{mygray}
    \multirow{-2}*{Type}&\multirow{-2}*{Method}   & PSNR $\uparrow$ & SSIM $\uparrow$ & PSNR $\uparrow$ & SSIM $\uparrow$ & PSNR $\uparrow$ & SSIM $\uparrow$  \\\hline
  & DesnowNet~\cite{liu2018desnownet}  & 20.13 & 0.81 &20.38 &0.84& 30.50 & 0.94 \\
    Desnowing& CycleGAN~\cite{zhu2017unpaired} &20.98 & 0.80 &20.21 &0.74& 26.81 & 0.89  \\
  Task& JSTASR~\cite{chen2020jstasr} & 27.96 &0.88 & 25.82 & 0.89 & 23.12 & 0.86 \\
  & DDMSNet~\cite{zhang2021deep} &   28.79 & 0.90 &27.03 &0.91& 30.76&0.91\\
   &HDCW-Net~\cite{chen2021all} &  29.06 &0.91 &27.78 &0.92 & 31.54 &0.95  \\
    \hline
  Adverse& All in One~\cite{li2020all} & 26.31 &0.87 &24.98 &0.88& 26.07&0.88  \\
   Weather& TransWeather~\cite{valanarasu2022transweather} &31.76 &0.93 & 28.29             & 0.92              & 31.82             & 0.93\\
 & TKL~\cite{chen2022learning} & 33.89 & 0.96 &30.82 &0.96& 34.37&0.95 \\
  \hline
  Universal& MPRNet~\cite{zamir2021multi} & 33.98 &0.97 &30.37  &0.96 & 33.87 &  0.95\\
  Image&Restormer~\cite{zamir2022restormer} &35.43 &0.97 &32.24  &0.96 & 34.67 & 0.95 \\
  Restoration & Maxim \cite{tu2022maxim}            & 38.08            & 0.98          &32.27           & 0.97     & 34.15             & 0.95 \\

  \hline
  \rowcolor{mygray}
  & Star-Net (Ours) & {\bbetter{25}{15}{\textbf{44.04}}{5.96}}&{\bbetter{25}{15}{\textbf{0.99}}{0.01}} & {\bbetter{25}{15}{\textbf{32.41}}{0.14}} &\textbf{0.97}& {\bbetter{25}{12}{\textbf{34.79}}{0.12}} &{\bbetter{25}{12}{\textbf{0.96}}{0.01}}  \\
  \hline\thickhline
 \end{tabular}
  \caption{Quantitative comparison of single image desnowing effect for Star-Net with several SOTA algorithms on the CSD, SRRS and Snow100K datasets.}
  \label{tab2}
  \end{table*}

\noindent\textbf{Datasets.} We conduct extensive experiments on three snow removal benchmark datasets, Snow100k \cite{liu2018desnownet}, SRRS \cite{chen2020jstasr}, and CSD \cite{chen2021all} to show the effectiveness of Star-Net in a single image desnowing task. 
The specific distribution of the three datasets is shown in Table~\ref{tab1}. Details of the datasets and the division of the dataset can be found in Appendix \textcolor{red}{A}.

\noindent\textbf{Settings. }During training, we use the Adam optimizer \cite{kingma2014adam} and set the base learning rate to 2e-5. 
The total training epochs are set to 210 and we halve the learning rate every 40 epochs to achieve dynamic adjustment for the learning rate. 
The input image size is uniformly cropped to 224x224. The default training batch size is 2. 
Our experiments are implemented on a single Nvidia RTX 3090. The specific experimental setup is shown in Appendix \textcolor{red}{B}.

\noindent\textbf{Comparison Methodology and Evaluation Metrics. }We select single image desnowing algorithms \cite{liu2018desnownet,chen2020jstasr,zhu2017unpaired,chen2021all,zhang2021deep}, adverse severe weather image recovery algorithms \cite{li2020all,valanarasu2022transweather,chen2022learning}, and image recovery algorithms \cite{zamir2021multi,zamir2022restormer,tu2022maxim} for comparison.  
For quantitative evaluation, we use two metrics, peak signal to noise ratio (PSNR) and structural similarity (SSIM), to evaluate the experimental results.

\subsection{Quantitative Evaluations}
The results of the quantitative evaluation of Star-Net on the CSD dataset are shown in Table~\ref{tab2}. 
As shown in Table~\ref{tab2}, Star-Net exhibits SOTA performance in the single image desnowing task and scores an amazing 44.04dB PSNR and 0.99 SSIM. 
Specifically, Star-Net gains 14.98dB in PSNR and 0.08 in SSIM compared to HDCW-Net \cite{chen2021all}. 
Compared to TKL \cite{chen2022learning}, Star-Net has a gain of 10.15dB in PSNR and 0.03 in SSIM. 
In addition, Star-Net improves PSNR by 5.06dB and SSIM by 0.01 compared to Maxim \cite{tu2022maxim}. 

In Table~\ref{tab2}, our Star-Net exhibits SOTA snow removal performance in the quantitative evaluation of the SRRS dataset. 
Formally, compared with DDMSNet \cite{zhang2021deep}, our Star-Net achieves a PSNR gain of 5.38dB and has an SSIM improvements of 0.06. 
Star-Net achieves a PSNR gain of 4.12dB and an SSIM gain of 0.05 compared to TransWeather \cite{valanarasu2022transweather}. 
In the SRRS dataset, Restormer \cite{zamir2022restormer} shows a PSNR of 32.24dB and SSIM of 0.96. Star-Net shows a PSNR of 32.41dB and SSIM of 0.97.

As shown in Table~\ref{tab2}, Star-Net performs best on the quantitative evaluation of the Snow100K dataset. 
Specially, we score a PSNR of 34.79dB and a SSIM of 0.96. This is demonstrated by the fact that we achieve a PSNR gain of 7.98dB and a SSIM improvement of 0.07 compared to Cycle-GAN \cite{zhu2017unpaired}. 
Compared to All in One \cite{li2020all}, Star-Net achieves a PSNR improvement of 8.72dB and SSIM gain of 0.08. 
Furthermore, Star-Net achieves a PSNR gain of 0.28dB and an SSIM improvement of 0.01 compared to MPRNet \cite{zamir2021multi}.

\subsection{Qualitative Evaluations}
Figure~\ref{fig6} shows the results of the qualitative comparison of Star-Net with several algorithms on CSD dataset. 
Although these methods achieve considerable snow removal on the CSD dataset, they still have some shortcomings. 
DesnowNet \cite{liu2018desnownet} and JSTASR \cite{chen2020jstasr} suffer from snow particle residual issue, and HDCW-Net \cite{chen2021all} suffers from image distortion. 
TransWeather \cite{valanarasu2022transweather} for severe weather also suffers from snow particle residuals and blurred images after snow removal. 
Maxim \cite{tu2022maxim} and our method exhibit better desnowing performance and the recovered image is closer to the ground truth. 
But there is a slight color difference in Maxim.

Figure~\ref{fig7} visualizes the snow removal results of Star-Net with several SOTA methods on the SRRS dataset. 
The comparison algorithm in Figure~\ref{fig7} shows excellent single image desnowing ability on the SRRS dataset. 
However, the rectangular border areas of DesnowNet \cite{liu2018desnownet}, JSTASR \cite{chen2020jstasr}, HDCW-Net \cite{chen2021all}, and TransWeather \cite{valanarasu2022transweather} desnowed images are still somewhat different from the ground truth. 
For Maxim \cite{tu2022maxim}, artifact issue appear on the SRRS dataset. 
Our Star-Net successfully removes all sizes of snow particles while reatining the closet clarity to the ground truth.

Figure~\ref{fig8} shows the snow removal effect of different algorithms on Snow100K. 
Though HDCW-Net \cite{chen2021all}, JSTASR \cite{chen2020jstasr} and TransWeather \cite{valanarasu2022transweather} remove most of the snow particles, there are still some snow particles with small scales remain. 
Maxim \cite{tu2022maxim} and DesnowNet \cite{liu2018desnownet} successfully removes snow particles from images. 
However, DesnowNet fails to recover the rectangular area in Figure~\ref{fig8} very well. And there are color biases in Maxim desnowing image compared to the ground truth. 
Compared with these SOTA methods, the clear images recovered by our Star-Net are much closer to the ground truth. See Appendix \textcolor{red}{D} for more visualization results.

\subsection{Ablation Study}
 \begin{table}[h]
  \centering
  \scriptsize
  \setlength{\belowcaptionskip}{-0.3cm}
  \begin{tabular}{c|ccc|c}
  \rowcolor{mygray}
  \hline\thickhline
   & \multicolumn{3}{c|}{\text { \textbf{Module} }} & \textbf { Metric }  \\\cline{2-4}
   \rowcolor{mygray}
  \multirow{-2}{*}{\textbf { Setting }}             & \text { SSC }     & \text{MIT}    & \text{DFM}  &\text { PSNR/SSIM } \\
  \hline 
  \text {S1}                                         &  $\times$         & \checked    &  \checked & 39.41 / 0.98 \\
  \text {S2}                                         &  \checked       & $\times$      &  \checked & 39.73 / 0.98 \\
  \text {S3}                                         &   \checked      &  \checked   & $\times$    &41.92 / 0.98 \\
  \text {S4}                                & \checked        &  \checked   &  \checked & 44.04 / 0.99\\
  \hline\thickhline
  \end{tabular}
  \caption{Ablation study of the proposed SSC, MIT, and DFM.}
  \label{tab3}
  \end{table}
To verify the effectiveness of the proposed algorithm, we perform ablation experiments on the three key designs of Star-Net. 
The settings in the ablation experiments are kept with the comparison experiments, and all the ablation experiments are performed on the CSD dataset.

\noindent\textbf{Effectiveness of SSC.} To demonstrate the enhancement of SSC for Star-Net single image desnowing ability, we replaced SSC with U-Net's \cite{ronneberger2015u} same layer skip connection for ablation studies. 
As shown in Table~\ref{tab3}, the PSNR has a 4.63 dB increase with the introduction of SSC, and the SSIM also has a 0.01 increase. 
This is because the SSC establishes information channels for different scale feature maps. 
These information channels can help Star-Net distinguish different sizes of snow particles and thus contribute to their removal. 
In addition, we show the qualitative and quantitative ablation results of SSC on Snow100K in Appendix \textcolor{red}{C}.

 \begin{table}[h]
  \centering
  \scriptsize
  \setlength{\belowcaptionskip}{-0.1cm}
  \resizebox{8.3cm}{!}{
  \begin{tabular}{c|cccc|c}
  \rowcolor{mygray}
  \hline\thickhline
   & \multicolumn{4}{c|}{\text { \textbf{Module} }} & \textbf { Metric }  \\\cline{2-6}
   \rowcolor{mygray}
  \multirow{-2}{*}{\textbf { Setting }}             & \text {MSAM}     & \text{MSAM to VA}    & \text{MSDC}  & \text{CGM}    &\text { PSNR/SSIM } \\
  \hline 
  \text {S5}                                         &  $\times$         & $\times$    &  \checked &  \checked   & 40.77 / 0.98 \\
  \text {S6}                                         &  $\times$       & \checked      &  \checked &  \checked   & 41.35 / 0.98 \\
  \text {S7}                                         &  \checked       & $\times$      &  $\times$ &  \checked   & 42.81 / 0.99 \\
  \text {S8}                                         &   \checked      &  $\times$   & \checked    &  $\times$   & 42.46 / 0.98 \\
  \text {S9}                                & \checked        &  $\times$   &  \checked &  \checked   & 44.04 / 0.99\\
  \hline\thickhline
  \end{tabular}}
  \caption{Ablation study of three core designs within MIT. MSAM stands for Multi Stage Attention Mechanisms. 
  MSAM to VA is the replacement of MASM with VA. MSDC represents Multi Scale Deep Convolution and CGM is Channel Gating Module.}
  \label{tab4}
  \end{table}

\noindent\textbf{Effectiveness of MIT.} We verify the effectiveness of the proposed MIT and the MIT internal multi stage attention mechanisms, multi scale deep convolution and convolutional gating network. 
For the ablation design of MIT, we replace MIT with the basic module of Restormer \cite{zamir2022restormer}. As shown in Table~\ref{tab3}, after removing MIT, the PSNR metric of Star-Net decreases by 4.31dB and the SSIM metric decreases by 0.01. 
For the three key modules within MIT, we also design ablation experiments for them. 
For the multi stage attention mechanisms, in addition to the ablation studies that removed it, we also design ablation experiments that replace it with the vanilla attention (VA) mechanism. For the multi scale deep convolution and convolutional gating network ablation design, we remove them separately. 
As can be seen from the experimental results in Table~\ref{tab4}, the removal of any key design within MIT leads to a decline in the evaluation metrics of Star-Net.

\begin{figure}[t]
\centering
\setlength{\abovecaptionskip}{0.2cm}
\setlength{\belowcaptionskip}{-0.4cm}
\includegraphics[width=1\linewidth]{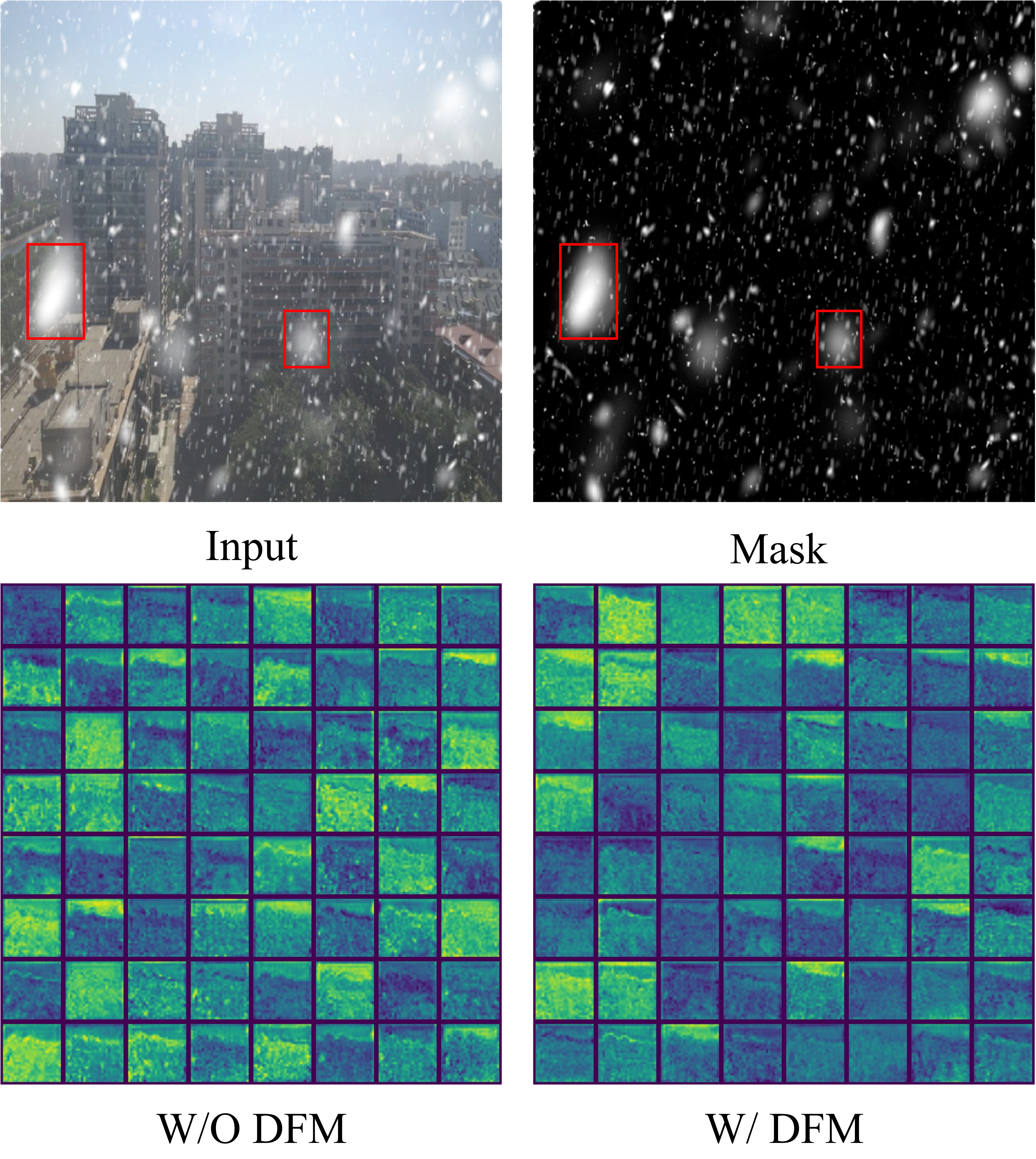}
\caption{Visualization of feature maps for DFM ablation studies. We visualize the feature maps for each channel without DFM and with DFM. 
For observation purposes, we only show the feature maps for 64 channels. Please focus on the rectangular area in the figure. 
When without DFM, snow particle residuals exist in the feature maps of some channels. 
When DFM is employed, snow particle residual is effectively removed. Please zoom in to see the details better.}
\label{fig9}
\end{figure}

\noindent\textbf{Effectiveness of DFM. }To prove the validity of the DFM, we remove it from Star-Net. The experimental results are shown in Table~\ref{tab3}. DFM helps our Star-Net to improve 2.12dB in PSNR and 0.01 in SSIM. 
This is because the DFM helps SSC filter out snow particle residuals contained in the spatial and channel level, resulting in more effective single-image desnowing. 
The feature visualization results in Figure~\ref{fig9} prove our hypothesis.

\section{Conclusion}
In this paper, we propose a novel end to end network for single image desnowing tasks called Star-Net. 
Specifically, Star-Net adopts a star type connection design SSC to introduce multi scale characteristics to remove snow particles with variable shapes. 
We design MIT that can explicitly model a variety of important image recovery features to address the image distortion issue related to the recovered image. 
In addition, by adding DFM to SSC for residual snow particle filtering on the spatial and channel domains, we further enhance the desnowing ability of Star-Net. 
Extensive experiments demonstrate that Star-Net achieves SOTA performance on the single image desnowing task.

\section*{Acknowledge} This work was supported by Public-welfare Technology Application Research of Zhejiang Province in China under Grant LGG22F020032, and Key Research and Development Project of Zhejiang Province in China under Grant 2021C03137.

\clearpage
\appendix

{\LARGE\noindent\textbf{Appendix}}

\vspace{0.5cm}

\section{Dataset Details}

\begin{figure}[h]
\centering
\includegraphics[width=1\linewidth]{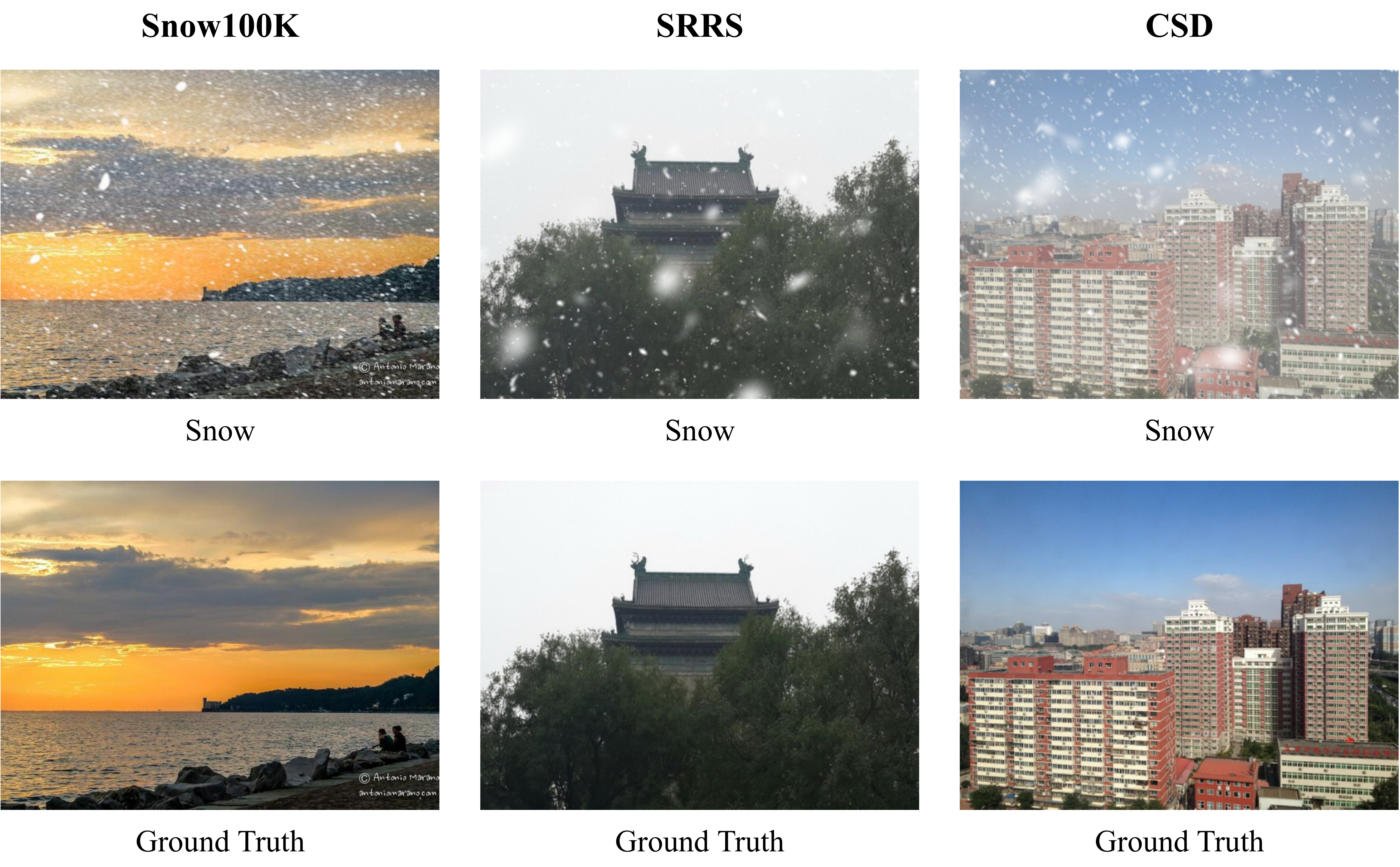}
\caption{Samples of three benchmark datasets Snow100K, SRRS, and CSD which are used in the experiments.}
\label{fig1}
\end{figure}

In the paper, we list the datasets used for the experiment. 
Next, we describe the details of each dataset and the division. Samples of the Snow100k \cite{liu2018desnownet}, SRRS\cite{chen2020jstasr}, and CSD \cite{chen2021all} datasets are shown in Figure~\ref{fig1}.

\noindent\textbf{Snow100K.} Snow100k dataset is synthesized from 100,000 snow images by setting different sizes, densities, shapes, transparency and motion trajectories of snow particles. 
Furthermore, the synthesized snow images are classified into three categories: Snow100K-S, Snow100K-M, and Snow100K-L for different snowfall amounts. 
Each image contains three parts: ground truth, synthetic snow images, and snow masks. 
In addition, the dataset retains the original aspect ratio of each image and normalizes the maximum boundary to 640 pixels. 

\noindent\textbf{Snow Removal in Realistic Scenario (SRRS).} SRRS dataset contains 15,000 synthetic snow images and another 1,000 snow images downloaded from the Internet for realistic scenes. 
At the same time, the haze dataset RESIDE \cite{li2018benchmarking} is used to synthesize images with masking effect, which makes the synthesized snow images closer to the real scenes. 

\noindent\textbf{Comprehensive Snow Dataset (CSD).} CSD consists of 10,000 synthetic snow images. During the image synthesis, 
the haze dataset RESIDE is applied to this dataset while following the synthesis method in the SRRS dataset to achieve the veiling effect. 
The dataset synthesizes different snowflakes and snow streaks for properties such as transparency, size and, position. 
A Gaussian blur is also applied to the snow particles to achieve a better simulation of real-world snow scenes. 

In the single image desnowing experiment of Snow100k, we randomly select 10,000 pairs of snow images from Snow100K-S, Snow100K-M, and Snow100K-L. 
Then we divide 8000 pairs of snow images out of 10000 pairs as the training set and 2000 pairs of snow images as the test set.
In the SRRS dataset, we also randomly selected 8000 pairs from 15000 pairs of snow images as the training set and 2000 pairs as the test set. 
For the CSD dataset, we follow the official dataset settings for training and testing.

\begin{table*}[t]
\centering
\scriptsize
\setlength\tabcolsep{2pt}
\renewcommand\arraystretch{1.1}
\begin{tabular}{c|c|c|c|c|c|c}
\hline\thickhline
\rowcolor{mygray}
& Multi Scale Attention & Criss Cross Attention & Window Attention  & Channel Attention & Fusion Channel Attention & CBAM \\
\hline
Layer 1           & D=256, H=2, SR=8, P=4 & C=16, $\text{Q}_\text{c}\text{=2}$, $\text{K}_\text{c}\text{=2}$, $\text{V}_\text{c}\text{=16}$, K=1 & WS=56, H=2, D=256, LN, P=4 & C=16, H=2, BIAS=True & C=64, H=2, BIAS=True & C=64, R=16, K=7 \\
Layer 2           & D=512, H=4, SR=8, P=4 & C=32, $\text{Q}_\text{c}\text{=4}$, $\text{K}_\text{c}\text{=4}$, $\text{V}_\text{c}\text{=32}$, K=1 & WS=28, H=4, D=512, LN, P=4 & C=32, H=4, BIAS=True & C=128, H=4, BIAS=True & C=128, R=16, K=7 \\
Layer 3           & D=1024, H=4, SR=4, P=4 & C=64, $\text{Q}_\text{c}\text{=8}$, $\text{K}_\text{c}\text{=8}$, $\text{V}_\text{c}\text{=64}$, K=1 & WS=14, H=4, D=1024, LN, P=4 & C=64, H=4, BIAS=True & C=256, H=4, BIAS=True & C=256, R=16, K=7 \\
Layer 4           & D=1024, H=2, SR=2, P=4 & C=64, $\text{Q}_\text{c}\text{=8}$, $\text{K}_\text{c}\text{=8}$, $\text{V}_\text{c}\text{=64}$, K=1 & WS=7, H=2, D=1024, LN, P=4 & C=64, H=2, BIAS=True & C=256, H=2, BIAS=True & C=256, R=16, K=7 \\
\hline\thickhline
\end{tabular}
\caption{Detailed architectural specifications of Multi Stage Attention Mechanisms.}
\label{tab1}
\end{table*}
\section{Experimental Details}

\begin{figure}[h]
\centering
\includegraphics[width=0.9\linewidth]{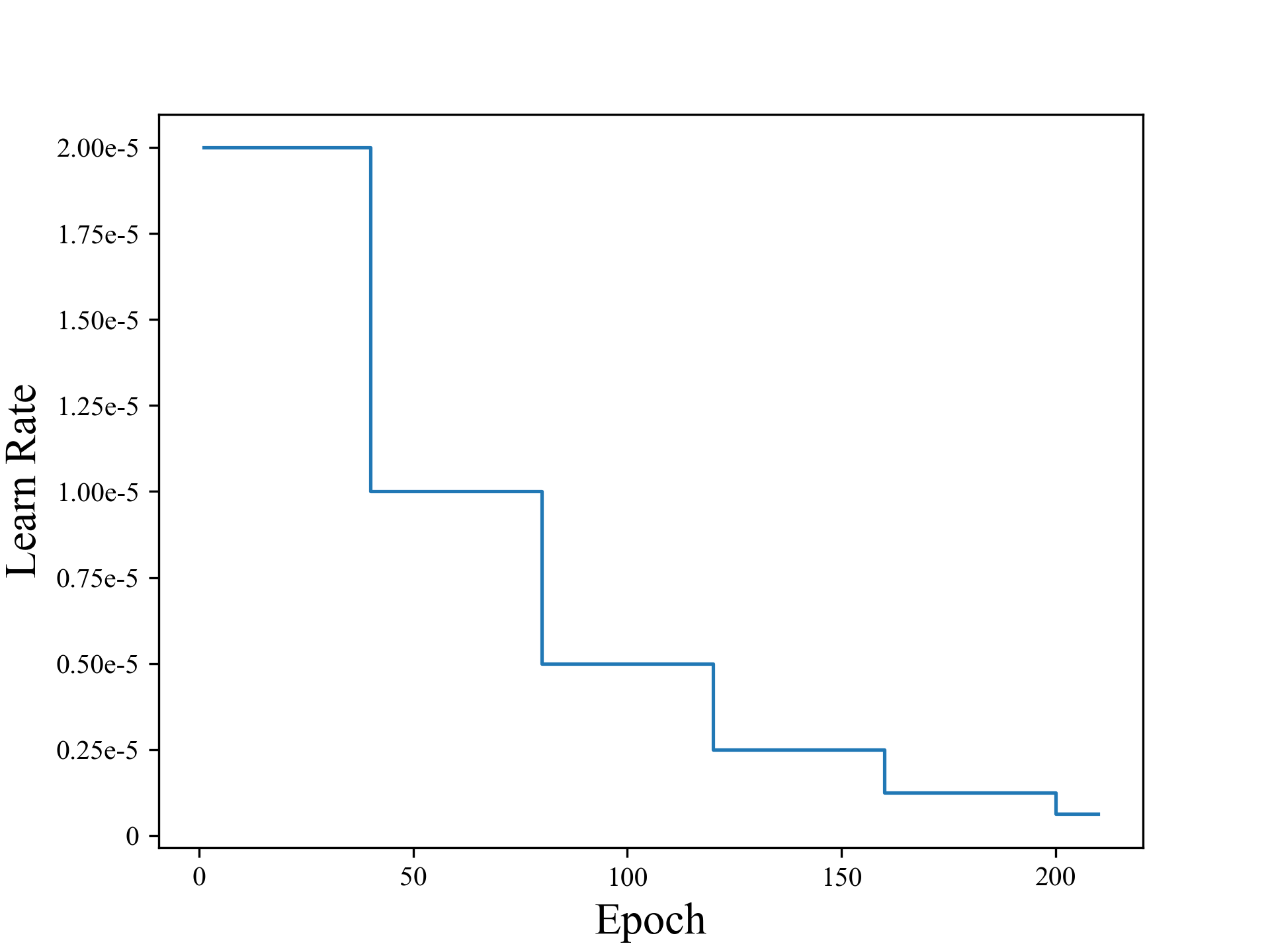}
\caption{Learning rate decay scheduler during training epochs.}
\label{fig2}
\end{figure}

\begin{table*}[h]
\centering
\footnotesize
\setlength\tabcolsep{1mm}
\renewcommand\arraystretch{1.1}
\begin{tabular}{c|c|c|c|c|c|c}
\hline\thickhline
\rowcolor{mygray}
& In Channel        & Out Channel       & Kernel Size       & Stride            & Padding           & Depth \\
\hline
Layer 1           & 64                & 128               & Low=3, Middle=5, High=7 & 1                 & Low=1, Middle=2, High=3 & 3 \\
Layer 2           & 128               & 256               & Low=3, Middle=5, High=7 & 1                 & Low=1, Middle=2, High=3 & 3 \\
Layer 3           & 256               & 512               & Low=3, Middle=5, High=7 & 1                 & Low=1, Middle=2, High=3 & 3 \\
Layer 4           & 256               & 512               & Low=3, Middle=5, High=7 & 1                 & Low=1, Middle=2, High=3 & 3 \\
\hline\thickhline
\end{tabular}
\caption{Detailed architectural specifications of Multi Scale Deep Convolution.}
\label{tab2}
\end{table*}

\noindent\textbf{Architecture details.} Table~\ref{tab1} shows the detailed architecture configurations of Multi Stage Attention Mechanisms in each layer of MIT. In Table~\ref{tab1}, D denotes the embedding dimension, and P indicates the patch size. 
H denotes the number of attention heads, and C indicates the feature map channel dimension. BIAS denotes whether bias is used in the convolution. 
In Multi Scale Attention mechanism, SR denotes the ratio of down-sampling. $\text{Q}_\text{c}$, $\text{K}_\text{c}$, and $\text{V}_\text{c}$ in Criss Cross Attention mechanism denote the number of channels of query, key, and value. K denotes the kernel size used in this convolutional layer. 
WS in Window Attention mechanism represents the window size of this layer. LN represents layer normalization. 
In the CBAM, R denotes the compression ratio of the spatial dimension and K denotes the size of the convolution kernel used for spatial attention in CBAM.
\begin{figure}[h]
\centering
\includegraphics[width=0.85\linewidth]{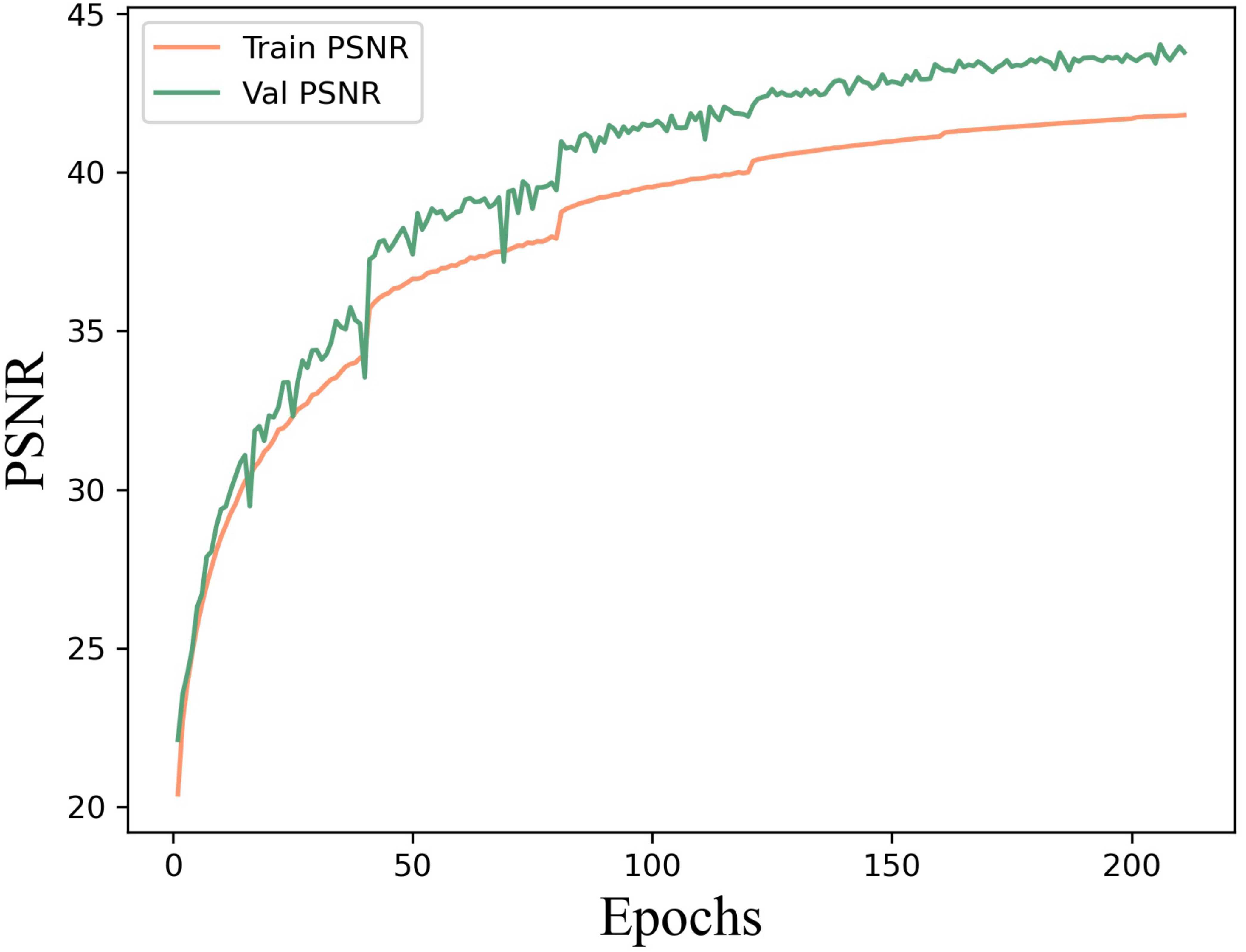}
\caption{Train PSNR and Val PSNR during training.}
\label{fig3}
\end{figure}
\begin{table}[h]
\centering
    \scriptsize
\begin{tabular}{c|c|c|c|c}
\hline\thickhline
\rowcolor{mygray}
& In Channel        & Out Channel       & Kernel Size       & Stride \\
\hline
Layer 1           & 64                & 64                & 3                 & 1 \\
Layer 2           & 128               & 128               & 3                 & 1 \\
Layer 3           & 256               & 256               & 3                 & 1 \\
Layer 4           & 256               & 256               & 3                 & 1 \\
\hline\thickhline
\end{tabular}
\caption{Detailed architectural specifications of Convolutional Gating Network.}
\label{tab3}
\end{table}
We show the specific details of Multi Scale Deep Convolution in Table~\ref{tab2}. 
In Channel indicates the input dimension of Multi Scale Deep Convolution in this layer. Out Channel represents the final output dimension of Multi Scale Deep Convolution.
Low, Middle, and High in Kernel Size denote the size of convolution kernel in different scale convolution operations. 
Stride denotes the step size of convolution. Low, Middle, and High in Padding indicate the number of padding pixels in different scale convolution operations. 
Depth means the depth of Multi Scale Deep Convolution.

Table~\ref{tab3} describes the specific configuration of the Convolutional Gating Network. In Channel and Out Channel represent the input and output dimensions of the gated convolution. 
Kernel Size and Stride represent the size and step size of the convolution kernel used in the gated convolution, correspondingly.

Table~\ref{tab4} illustrates the specific configuration of each hyperparameter in DFM. Patch Size denotes the size of the patch in the Spatial Self-Attention Module. 
Channel indicates the number of channels in the DFM feature map for each layer. Embed Dim denotes the embedding dimension of the DFM Spatial Self-Attentive Module. Num Heads denotes the number of attention heads of the Spatial Self-Attention Module. Expansion Factor denotes the multiplier of MLP hidden layer dimension expansion. 
$\text{K}_\text{c}$ represent the size of the convolution kernel, and $\text{K}_\text{n}$ denotes the size of the convolutional kernel used for the DFM-to-DFM branch of the different layers of SSC. Groups represent the number of channel groups in DFM Channel Gating Module. 

\begin{figure}[t]
\centering
\includegraphics[width=0.88\linewidth]{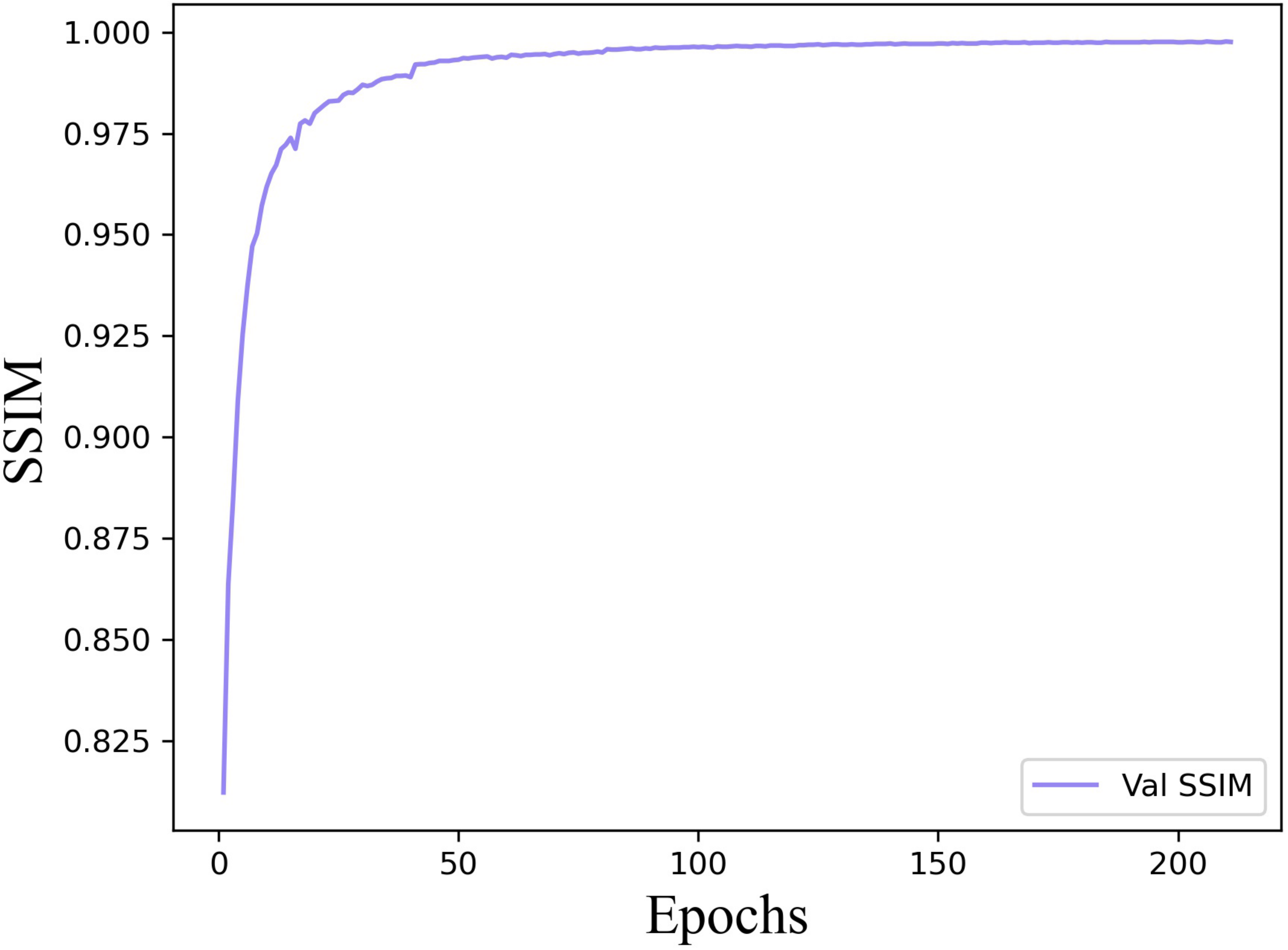}
\caption{Val SSIM during training.}
\label{fig4}
\end{figure}

\begin{table*}[h]
\centering
    \scriptsize
\begin{tabular}{c|c|c|c|c|c|c|c}
\hline\thickhline
\rowcolor{mygray}
& Patch Size        & Channel           & Embed Dim         & Num Heads         & Expansion Factor  & Kernel Size       & Groups \\
\hline
Layer 1           & 4                 & 64                & 1024              & 2                 & 1                 & $\text{K}_\text{c}\text{=3}$, $\text{K}_\text{n}\text{=7}$        & 128 \\
Layer 2           & 4                 & 128               & 2048              & 4                 & 1                 & $\text{K}_\text{c}\text{=3}$, $\text{K}_\text{n}\text{=5}$       & 256 \\
Layer 3           & 4                 & 256               & 4096              & 4                 & 1                 &$\text{K}_\text{c}\text{=3}$, $\text{K}_\text{n}\text{=3}$     & 512 \\
Layer 4           & 4                 & 256               & 4096              & 2                 & 1                 & $\text{K}_\text{c}\text{=3}$, -         & 512 \\
\hline\thickhline
\end{tabular}
\caption{Detailed architectural specifications of Degradation Filter Module.}
\label{tab4}
\end{table*}

\noindent\textbf{More details about implementation.} We scale the image to $224 \times 224$ pixels. We set the exponential decay rate $\beta_1$ and $\beta_2$ of the Adam optimizer \cite{kingma2014adam} to 0.9 and 0.999, 
and the weight decay to 1e-8. In the perceptual loss, we calculate it with the features extracted from layers 3, 8, and 15 of the VGG16 network \cite{simonyan2014very}. 
In smooth L1-loss \cite{girshick2015fast}, we set the $\beta$ to 1. 
Figure~\ref{fig2} shows the dynamic evolution of the learning rate during Star-Net training on the CSD dataset. In addition, we demonstrate the fluctuations of PSNR and SSIM in Figure~\ref{fig3} and Figure~\ref{fig4}.
\begin{figure}[h]
\centering
\includegraphics[width=1\linewidth]{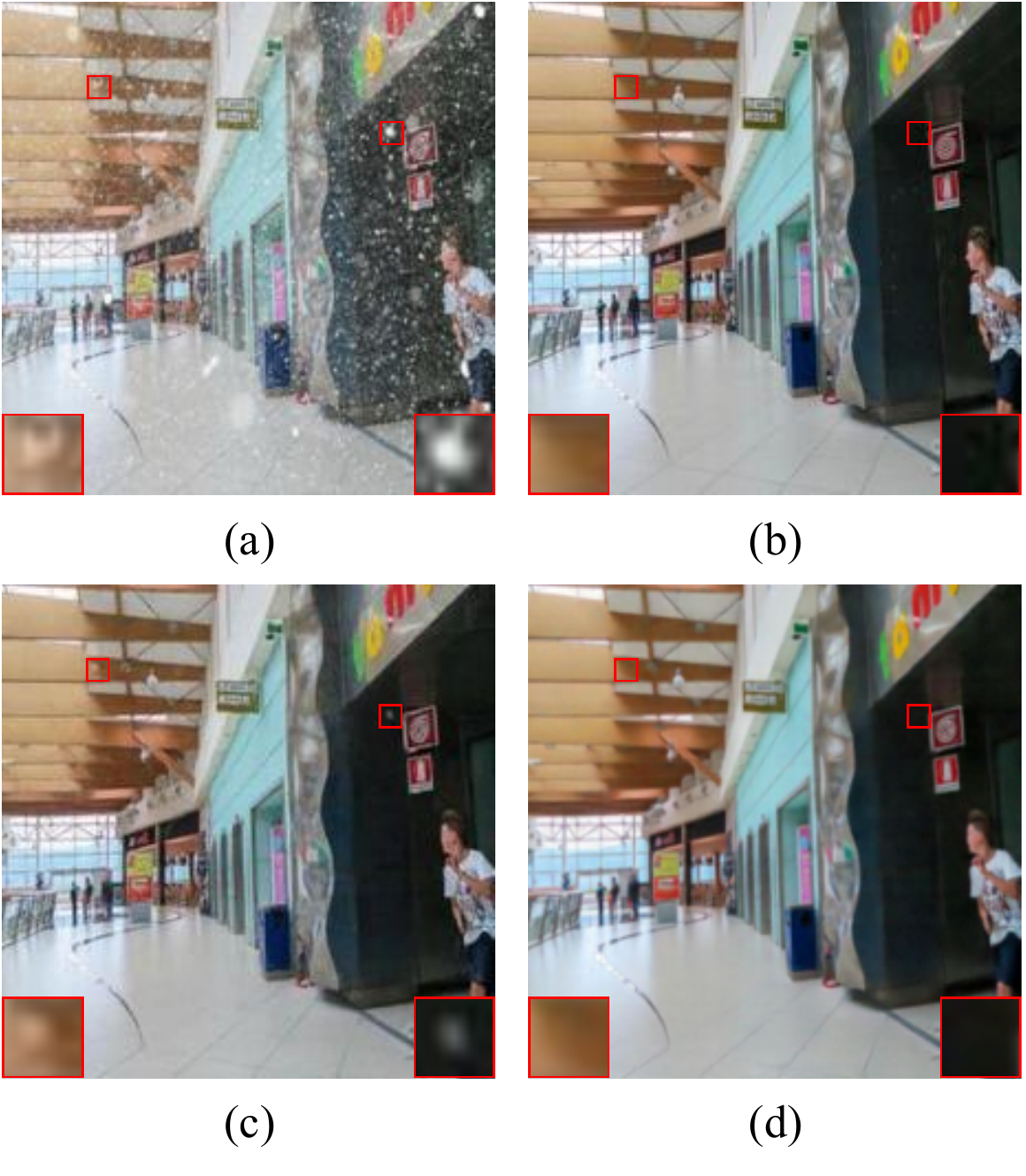}
\caption{Visualization of the SSC ablation experiment on Snow100K.(a) Input Image. (b) Ground Truth. (c) W/O SSC. (d) W/ SSC. Please zoom in to see the details better.}
\label{fig5}
\end{figure}

\section{Experimental Supplement}

In this section, we complement the SSC ablation experiments in Section \textcolor{red}{4.5} of the main text. 
Specifically, we ablate the SSC in Snow100K following the same settings as described in the main text. 
And the ablation results of SSC on Snow100K are shown in Table~\ref{tab5} and Figure~\ref{fig5}, respectively. As shown in Table~\ref{tab5}, the SSC ablation experiments on Snow100K exhibited the same results on CSD. 
As shown in Figure~\ref{fig5}, when Star-Net uses the same-layer skip connection of U-Net\cite{ronneberger2015u}, it removes most snow particles from the input image by MIT and DFM. 
However, there are still minor small-sized snow particles remaining in the rectangular area being plotted. After the introduction of SSC, all sizes of snow particles in the input image are completely removed. 
The above results illustrate that the structural design of the multi scale skip connection makes the model more effective in snow removal in scenarios with different sizes of snow particles compared to the single-scale skip connection.
\begin{table}[h]
\centering
\scriptsize
\begin{tabular}{c|c|c}
\rowcolor{mygray}
\hline\thickhline
& \multicolumn{1}{c|}{\text { \textbf{Module} }} & \textbf { Metric }  \\\cline{2-3}
\rowcolor{mygray}
\multirow{-2}{*}{\textbf { Setting }}             & \text { SSC }      &\text { PSNR/SSIM } \\
\hline 
\text {S1}                                         &  $\times$         & 31.58 / 0.95 \\
\text {S2}                                         &  \checked         & 34.79 / 0.96 \\
\hline\thickhline
\end{tabular}
\caption{Ablation study of the proposed SSC, MIT, and DFM.}
\label{tab5}
\end{table}

\section{Additional Visual Comparisons}
We present more visualization comparison results of Star-Net with other algorithms on CSD, SRRS, and Snow100k datasets in Figure~\ref{fig6}, Figure~\ref{fig7}, and Figure~\ref{fig8}. 
Compared with other algorithms, our Star-Net is more effective in removing snow particles in different snow scenes. 
And our recovered images are closer to the ground truth.
\clearpage

\begin{figure*}[h]
\centering
\includegraphics[width=1\linewidth]{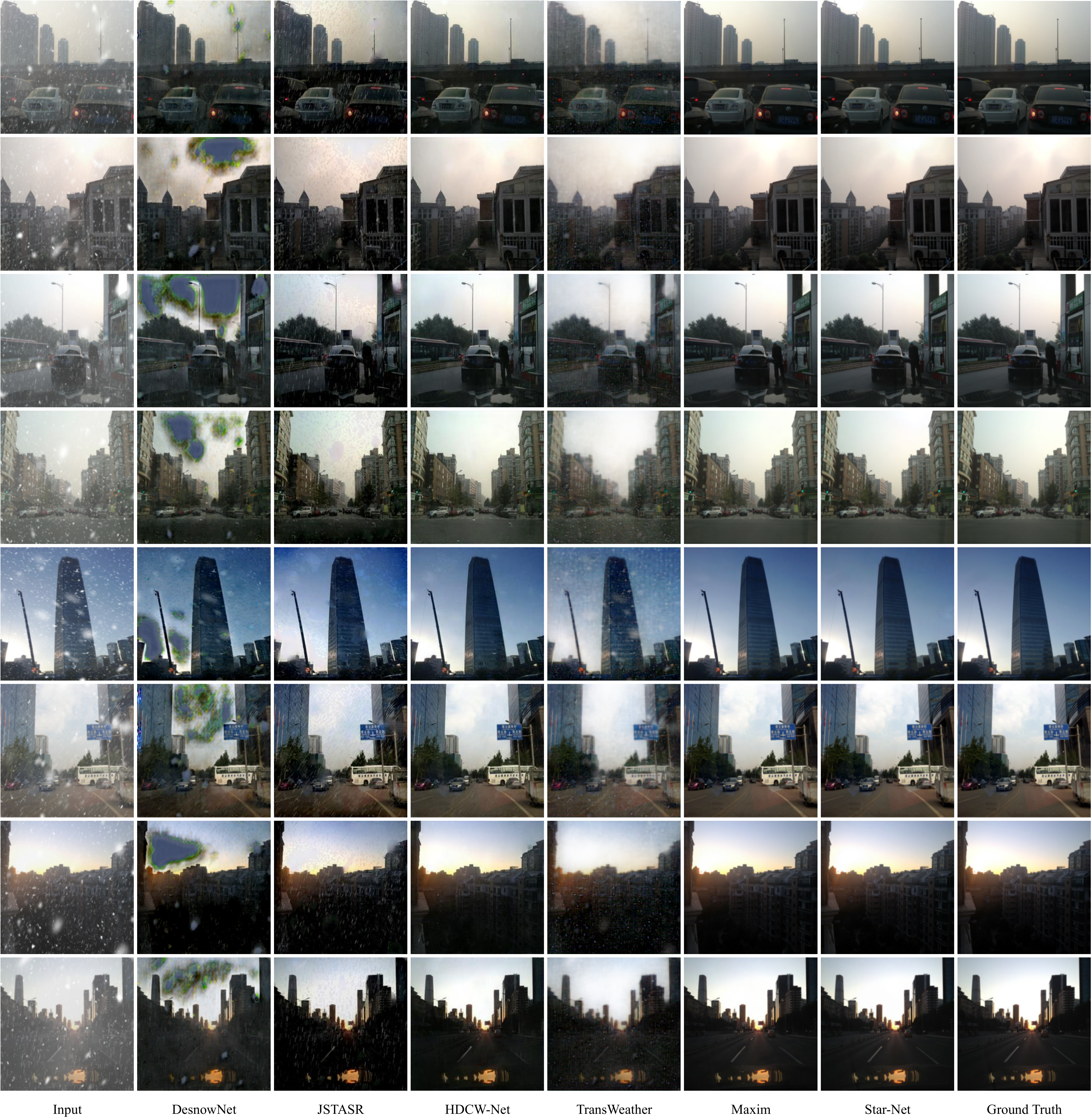}
\caption{Visualization comparison of DesnowNet \cite{liu2018desnownet}, JSTAST \cite{chen2020jstasr}, HDCW-Net \cite{chen2021all}, 
TransWeather \cite{valanarasu2022transweather}, Maxim \cite{tu2022maxim} and proposed method on the CSD dataset.}
\label{fig6}
\end{figure*}

\begin{figure*}[h]
\centering
\includegraphics[width=1\linewidth]{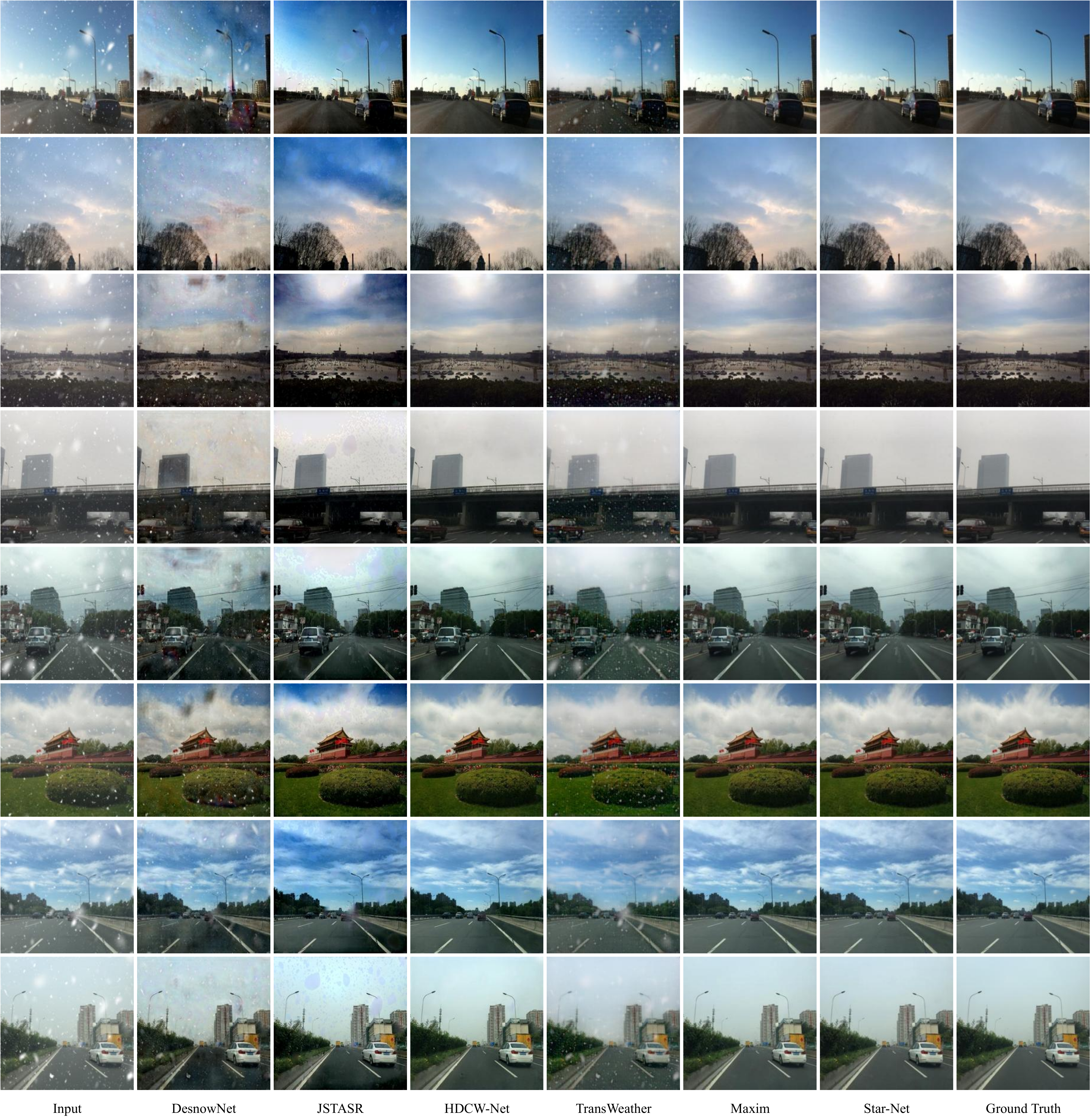}
\caption{Visualization comparison of DesnowNet \cite{liu2018desnownet}, JSTAST \cite{chen2020jstasr}, HDCW-Net \cite{chen2021all}, 
TransWeather \cite{valanarasu2022transweather}, Maxim \cite{tu2022maxim} and proposed method on the SRRS dataset.}
\label{fig7}
\end{figure*}

\begin{figure*}[h]
\centering
\includegraphics[width=1\linewidth]{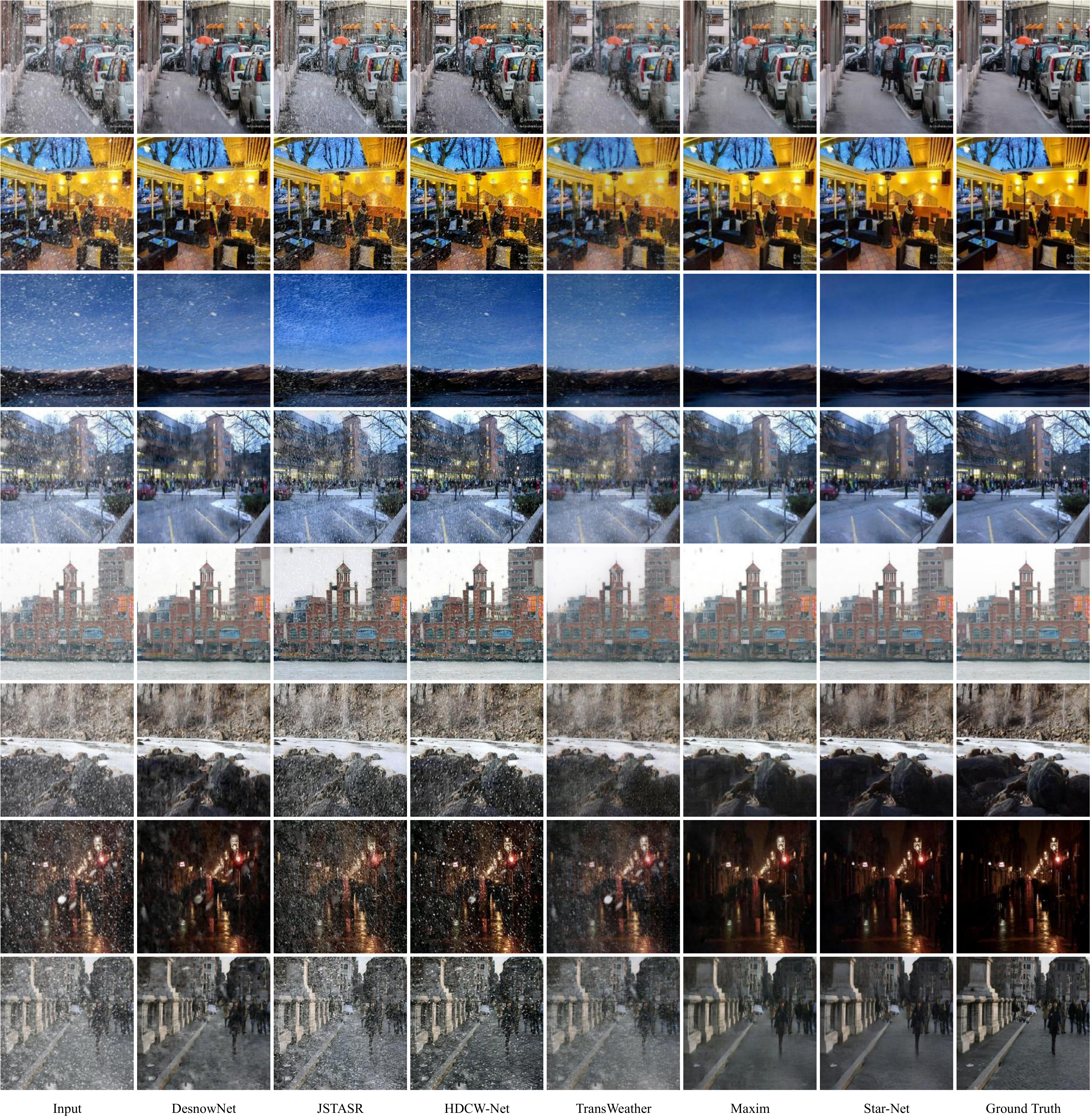}
\caption{Visualization comparison of DesnowNet \cite{liu2018desnownet}, JSTAST \cite{chen2020jstasr}, HDCW-Net \cite{chen2021all}, 
TransWeather \cite{valanarasu2022transweather}, Maxim \cite{tu2022maxim} and proposed method on the Snow100K dataset.}
\label{fig8}
\end{figure*}

\clearpage
{\small
\bibliographystyle{ieee_fullname}
\bibliography{egbib}
}

\end{document}